\documentclass[lettersize, journal]{IEEEtran}

\usepackage{amsmath,amssymb,amsfonts}
\usepackage{algorithmic}
\usepackage{graphicx}
\usepackage{textcomp}
\usepackage{booktabs}
\usepackage{xcolor}
\usepackage{verbatim}
\usepackage{hyperref}
\usepackage{caption}
\usepackage{subcaption}
\usepackage{tikz}
\usepackage{balance}

\def\BibTeX{{\rm B\kern-.05em{\sc i\kern-.025em b}\kern-.08em
    T\kern-.1667em\lower.7ex\hbox{E}\kern-.125emX}}

\makeatletter
\newcommand{\linebreakand}{%
  \end{@IEEEauthorhalign}
  \hfill\mbox{}\par
  \mbox{}\hfill\begin{@IEEEauthorhalign}
}
\makeatother

\usepackage{soul}
\definecolor{lightgreen}{RGB}{218, 243, 218} 
\definecolor{lightred}{RGB}{243, 223, 218} 
\definecolor{lightgrey}{RGB}{233, 233, 233} 
\newcommand{\hlc}[2][lightgreen]{{%
    \colorlet{foo}{#1}%
    \sethlcolor{foo}\hl{#2}}%
}

\usepackage[style=numeric,sorting=none,backend=biber]{biblatex}
\begin{document}

\title{Emotional regulation improves deep learning-based image classification}

\author{\IEEEauthorblockN{Riccardo Emanuele Landi\thanks{Riccardo Emanuele Landi is with Mare Group, Salerno, Italy (e-mail: riccardo.landi@maregroup.it).}, Jo\~{a}o M. F. Rodrigues\thanks{Jo\~{a}o M. F. Rodrigues is with NOVA LINCS and Institute of Engineering (ISE), University of Algarve, Faro, Portugal (e-mail: jrodrig@ualg.pt).}, and Marta Chinnici\thanks{Marta Chinnici is with the Department of Energy Technologies and Renewable Sources, ENEA Casaccia Research Center, Rome, Italy (e-mail: marta.chinnici@enea.it).}}}

\markboth{}{}

\maketitle
\thispagestyle{empty}

\begin{abstract}
Emotion significantly influences cognition, enhancing memory and learning under certain conditions. Drawing on this principle, emotion-augmented deep learning investigates how affective states can improve neural network architectures and learning paradigms, achieving better generalization than non-emotional models. However, existing methods often rely solely on objective neurophysiological factors, neglecting the role of subjectivity in emotion. To bridge this gap, the present study introduces Emotional Regulation, a novel framework for modeling emotion in deep learning through artificial subjective experience. The method employs pre-training based on affective stimuli, balancing non-emotional and emotionally-influenced responses in downstream task optimization. Extensive experimentation was conducted in image classification, pre-training ResNet and ViT architectures on four emotional datasets, using CIFAR-10 and -100 as target benchmarks. Results reveal improvements over the aforementioned backbones, providing evidence of Emotional Regulation as a promising method for defining emotion-augmented deep learning through artificial subjective experience. Furthermore, the proposed approach overcomes the related work in image classification based on CIFAR, revealing Emotional Regulation as the new state-of-the-art in emotion-augmented deep learning for large-scale vision datasets. The study also enforces evidence of the impact of affective states in improving machine learning tasks' optimization, encouraging further investigation on emotion-inspired architectures.
\end{abstract}


\section{Introduction}

\IEEEPARstart{E}{motion} is a significant factor in human cognition, as internal emotional states affect perception, attention, learning, memory, and other relevant processes \cite{keating2023, zadra2011, becker2011, vanhoff2011, yin2023, tyng2017, congleton2020, brosch2013impact, bechara2004role, li2020role}. Affective states impact learning and memory, as humans tend to process emotion-evoking visual stimuli with enhanced salience \cite{tyng2017, congleton2020, yin2023, pan2025positive}. Building on this, artificial intelligence systems increasingly draw inspiration from human cognitive processes and behavior, developing advanced computational models to enhance machine learning architectures \cite{iovane2023smart, assunccao2022overview, brauwers2021general, van2020brain, moerland2018emotion, jacobs2014emergent, landi2023cognitivenet}. In this context, \textit{Emotion-augmented Machine Learning} \cite{stromfelt2017, assunccao2022overview} defines the set of methods introducing emotion in the artificial learning process to improve generalization. Studies have demonstrated the effectiveness of employing emotion as an enhancing factor, encouraging the incorporation of affective states into task optimization. Deep learning architectures inspired by the structure of the emotional brain have been proposed, as well as approaches involving custom definitions of backpropagation \cite{lotfi2012, lotfi2013, parvinizadeh2022, embp2008, zare2022, thenius2013, khashman2011credit}. However, these architectures define emotion primarily through its neurophysiological characteristics, in terms of neural structures and hormonal processes, neglecting the role of subjectivity, which is a significant factor influencing the emotional response. This latter aspect was recently investigated through the \textit{Emotional Regulation} approach, which improves generalization through the balancing of non-emotional and emotionally-influenced predictions \cite{landi2024}. The pioneering study reported improvements in the task of image classification, revealing preliminary evidence of the approach's validity; however, the method was not implemented in an end-to-end learning architecture and was not validated through different sets of emotional stimuli, limiting the robustness of the results and the generalization of the methodology. 

Given these premises, the present study aims to propose Emotional Regulation as a promising approach for modeling emotion in deep learning, proving its effectiveness in optimizing fundamental tasks, such as image classification. To achieve this objective, the original framework was extended to end-to-end learning for enhanced optimization, employing convolutional neural networks (ResNet, in the present work) \cite{he2016} and vision transformer (ViT) \cite{dosovitskiy2020} as backbone architectures. Experimentation with different sets of emotional stimuli was conducted to ensure a robust basis for proving the generalization of the results. The main contributions of the paper consist of: (i) providing further evidence of the effectiveness of emotion-inspired architectures, proposing a novel approach for defining affective states in deep learning; (ii) bridging the gap in current state-of-the-art architectures, which do not consider subjective factors in emotion, proposing Emotional Regulation as a significant method for modeling affective states; (iii) extending Emotional Regulation to end-to-end learning, pre-training backbones on different sets of emotional stimuli, to prove the effectiveness of the framework; (iv) providing evidence that emotion improves deep learning-based image classification, comparing emotional and non-emotional models on benchmarking datasets; (v) proposing \textit{emotional pre-training} (i.e., pre-training based on a history of artificial experiences) as a promising approach for studying emotion in artificial intelligence; (vi) proving that Emotional Regulation outperforms existing solutions in emotion-augmented deep learning on large scale image classification datasets.

The present work does not claim to present a model of human cognition, nor to equate the results obtained in the context of deep learning with empirical evidence from studies involving human beings. The multidisciplinary contribution of the work lies in investigating emotion as a means to enhance deep learning, drawing inspiration from findings in cognitive science.

Section \ref{emotion_augmented_deep_learning} reviews the essential scientific evidence underpinning the present investigation, presenting how emotion improves human learning and memory in visual stimuli processing. The above Section also provides an overview of state-of-the-art methods in emotion-augmented deep learning, discussing their characteristics and limitations. Subsequently, Sections \ref{emotional_regulation} and \ref{experimental_setup} define the Emotional Regulation framework with the related experimental setup. The study concludes with the presentation and discussion of the results, indicating implications and future directions. 

\section{Emotion-augmented deep learning}
\label{emotion_augmented_deep_learning}

\subsection{Emotion improves learning and memory}
\label{emotion_improves_memory_and_learning}

Emotional activity is significantly associated with the limbic system, involving the sensory cortex, prefrontal cortex, orbitofrontal cortex, hypothalamus, hippocampus, and thalamus \cite{stromfelt2017, assunccao2022overview}. Emotion plays a pivotal role in improving memory and learning of stimuli \cite{tyng2017}. In this scenario of scientific evidence, Congleton \& Berntsen \cite{congleton2020} employed a first-person perspective video simulation to examine the impact of emotional valence on memory. They assessed participants' recall accuracy for both central and peripheral details in the context of positive and negative stimuli, revealing superior retention of scene details in the negative-valence condition. In a subsequent meta-analysis, Yin et al. \cite{yin2023} proved that emotional stimuli significantly influence both judgments of learning (JOLs) and memory, revealing improvements for both positive and negative valence, proving the highest impact of visual stimuli compared to verbal content.

The impact of emotional states on learning was also investigated by Kremer et al. \cite{kremer2019}. They conducted a study in which negative emotions were elicited in medical residents to observe the related effect on learning. Also in this case, the authors assumed visual stimuli as the sensory channel for evoking emotions. They found that students who watched emotional videos performed worse on a test after studying a reference subject. Negative emotions led to a decrease in the time invested in learning and the acquisition of knowledge. However, unlike the other aforementioned studies, in this case, emotion was evoked by a stimulus external to the process for which the impact on learning was intended to be verified. In fact, evidence shows that a visual stimulus has a greater impact on learning and memory when it represents the direct cause of emotion elicitation. Furthermore, another study provided evidence of an improvement in memory in the condition of positive emotional states. In particular, Pan et al. \cite{pan2025positive} found that a neutral visual stimulus is significantly evoked when a positive emotion-eliciting image is presented subsequently in the same context. In the present study, it is intended to focus on cognitive processes involving the enhancement of memory and learning of stimuli that directly elicit emotional states, without considering other conditioning instances.

\subsection{Emotion as a deep learning enhancer}
\label{emotion_as_a_deep_learning_enhancer}

The above evidence allows to assume that emotion-evoking visual stimuli are learnt with better accuracy by a human, motivating investigations of a similar effect in deep learning models. In this context, emotion was introduced in artificial learning to improve deep neural networks' performance. A characteristic approach in this scenario is represented by \textit{brain emotional learning} (BEL) \cite{lotfi2012, lotfi2013}, which involves the design of neural networks based on the anatomical structure of the emotional brain. Following this path, Shahid \& Singh \cite{shahid2020} employed an architecture considering the amygdala, sensory cortex, orbitofrontal cortex, and thalamus, revealing improvements in particle swarm optimization for coronary artery disease diagnosis. Improvements were also found by investigating the same approach in forecasting tasks \cite{zamirpour2018, parvinizadeh2022, suthasinee2024}, particularly implementing recurrent neural networks by introducing weighted adjustable connections between orbitofrontal cortex and amygdala \cite{kumar2005}. Furthermore, architectures focusing on the functioning of the prefrontal cortex were proposed, integrating visual stimuli processing with context information. In particular, Xu et al. \cite{xu2021} proposed a memristive circuit performing multitask digit classification, assuming a task identifier as the context. The results provided better performance compared with similar methods.

Another significant approach for modeling emotion in neural networks is based on \textit{emotional backpropagation} (EmBP) \cite{embp2008, embp2009}. Within this paradigm, artificial emotion consists of hormonal glands affecting neurons, employing a backpropagation process involving factors related to anxiety and confidence. Along this path, Thenius, Zahadat \& Schmickl \cite{thenius2013} simulated the neuromodulatory hormonal system, proving the effectiveness of the approach. Emotional backpropagation was also investigated for estimating water quality \cite{abba2022}, revealing enhancements in predicting a target index. Being dopamine a neurotransmitter significantly influencing learning, Zare et al. \cite{zare2022} proposed an approach employing the regulation of the learning rate based on positive and negative emotional states. They modeled the human dopamine fluctuations in emotional neural networks, overcoming traditional methods in addressing face recognition. The approach assumed anxiety and confidence factors in the learning process.

The above approaches do not consider the subjective nature of emotion. Even though there exist objective neurophysiological processes influencing emotion and learning, subjective experience needs to be investigated to define a holistic framework \cite{brown2020coherence,zhang2025neurofunctional}. In this scenario, a suitable path is tracked by the attempt to introduce an emotional history in the learning process. In fact, an emotional history can be defined as a set of emotional experiences influencing reactions to stimuli. For instance, a subject can react to a given stimulus as a consequence of personality -- subjective nature, in general -- or past experiences related to that sensory instance. Emotion can therefore be introduced by modeling the above process to integrate neural networks with artificial subjective experience, designing models that estimate emotional reactions to stimuli; thus, the emotional history would consist of the dataset employed for training an emotional model \cite{wang2025enhancing, zhou2025improved, liang2024, luo2025, zhou2025object}.

The above gaps are addressed in the present study by investigating Emotional Regulation \cite{landi2024} as a model of emotion. The approach involves pre-training an emotional encoder on a dataset of stimulus-emotion instances, balancing non-emotional and emotionally-influenced predictions in the function of the current affective state. This approach allows for introducing subjective emotional reactions in the learning process, overcoming the definition of emotion-augmented architectures strictly focused on neurophysiological factors. The framework also allows for easily conducting an investigation on the significance of emotion (in particular, of subjective emotional outcome) in improving image classification, as the method was originally designed to address visual stimuli processing. An extensive experimentation of Emotional Regulation on this fundamental task allows to achieve strong generalized evidence of the impact of emotion in deep learning-based vision.

Regulation mechanisms are of particular interest in artificial emotion modeling; in this context, a valuable study was conducted by Man \& Damasio \cite{man2020truth, man2019homeostasis}, through their investigation of homeostatic self-regulation in artificial neural networks. Drawing inspiration from self-preservation processes in living organisms, the authors proposed the design of machines that take into account the preservation of an optimal state. The present study aims to enhance the above hypothesis by extending the concept of homeostasis to subjective emotional experience. Along this path, the Emotional Regulation framework represents a suitable approach for defining a process of affective state regulation, with the aim of preserving an optimal learning condition in artificial neural networks.

Although several studies have been conducted in the context of emotion modeling in deep learning, this research topic remains underexplored. The present work aims to encourage investigation of the subjective aspects of emotion, introducing a novel perspective through the framework of Emotional Regulation.

\section{Emotional Regulation}
\label{emotional_regulation} 

\begin{figure*}
\centering
\begin{subfigure}[t]{0.70\textwidth}
    \centering
    \begin{tikzpicture}
        \node[anchor=south west, inner sep=0] (image) at (0,0) {%
            \includegraphics[width=\textwidth]{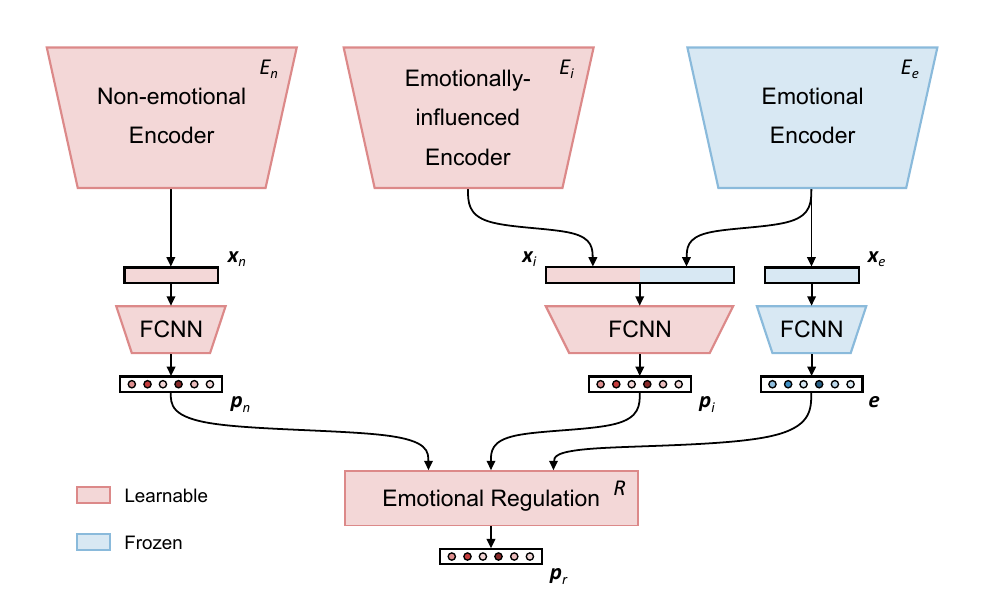}};
        \node[anchor=north west, inner sep=2pt] at (image.north west) {\textbf{a}};
    \end{tikzpicture}
    \label{fig:emotional_regulation}
\end{subfigure}

\vspace{1em}

\begin{subfigure}[t]{0.70\textwidth}
    \centering
    \begin{tikzpicture}
        \node[anchor=south west, inner sep=0] (image) at (0,0) {%
            \includegraphics[width=\textwidth]{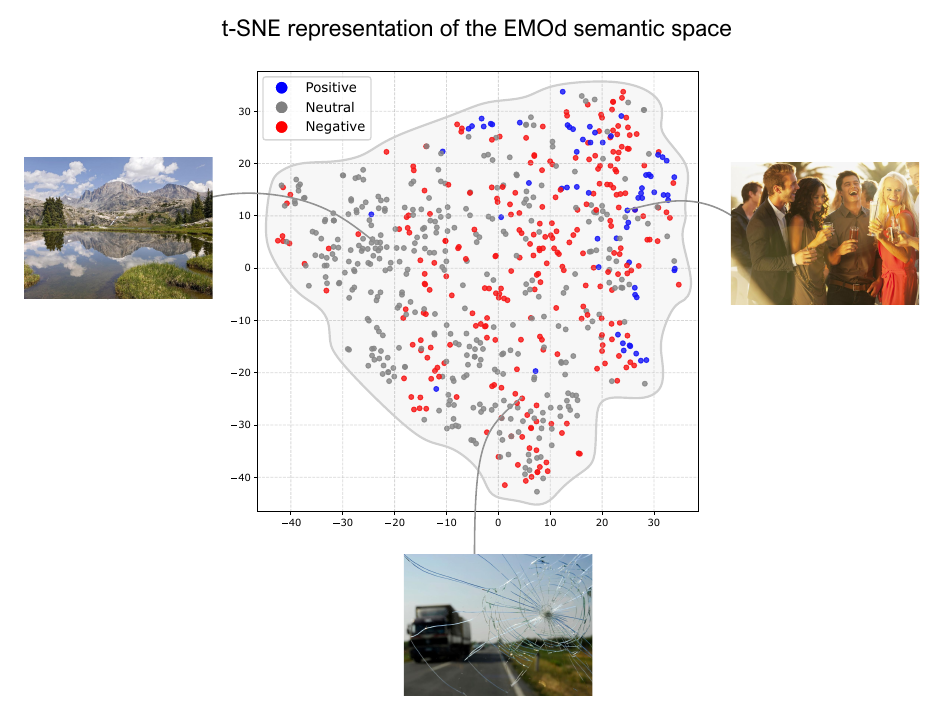}};
        \node[anchor=north west, inner sep=2pt] at (image.north west) {\textbf{b}};
    \end{tikzpicture}
    \label{fig:emot_inf_scatter_imgs}
\end{subfigure}

\caption{(a) Schema of the Emotional Regulation framework. (b) A t-SNE representation of the EMOd semantic space based on CLIP text embeddings; the three images in the example are projected in a semantic space associated with categorical valence, i.e., neutral, negative, and positive classes of emotion, respectively.}
\label{fig:combined_emotion}
\end{figure*}

Emotion provides semantics associated with internal and external stimuli, enabling the integration of the perceptual features of world instances with deeper meanings. A visual stimulus is not described solely by its chromatic, geometric, spatial, or linguistic properties, but also by the emotional experience associated with that stimulus. The Emotional Regulation framework combines emotional and non-emotional predictions in the function of the elicited affective state, pre-training an emotional architecture based on a set of artificial subjective experiences. As shown in Fig. \ref{fig:combined_emotion}(a), the proposed approach is composed of three encoders, i.e., the Non-emotional, Emotionally-influenced, and Emotional Encoder, each coupled with a fully-connected neural network (FCNN). The Non-emotional Encoder ($E_{n}$) represents the original backbone and produces non-emotional representations. The second encoder, i.e., the Emotionally-influenced ($E_{i}$), learns embeddings that consider emotion through feature concatenation involving the Emotional Encoder ($E_{e}$). The non-emotional and emotionally-influenced models are learnable, while the Emotional Encoder, together with the related fully-connected layers, is frozen, pre-trained on a given emotional history.

The Non-emotional and Emotionally-influenced Encoders are trained on a downstream task, learning representations $\textbf{x}_{n} \in \mathbb{R}^{B \times F}$ and $\textbf{x}_{i} \in \mathbb{R}^{B \times 2F}$, providing probability vectors $\textbf{p}_{n} \in {[0,1]}^{B \times C}$ and $\textbf{p}_{i} \in {[0,1]}^{B \times C}$, respectively, where $B$ is the batch size, $F$ are the feature dimensions, and $C$ is the cardinality of classes. The Emotional Encoder is trained on a set of emotional stimuli, learning representations $\textbf{x}_{e} \in \mathbb{R}^{B \times F}$ associated with the emotional spectrum $\textbf{e} \in {[0,1]}^{B \times E}$, with $B$ the batch size and $E$ the cardinality of the considered emotional categories. Output instances $\textbf{p}_{n}$, $\textbf{p}_{i}$, and $\textbf{e}$ are computed through softmax activation. In the framework, emotion is assumed to be a vector of probabilities associated with classes. This representation is projected in the valence space coherently with the class of maximum probability. Emotional categories can be described by any model of emotion that assumes probability vectors as target output.

The last element of the model is represented by the Emotional Regulation process ($R$), which weights non-emotional or emotionally-influenced predictions by evaluating the emotional outcome. Formally,

\begin{equation}
    \textbf{p}_{r} = R(\textbf{p}_{n}, \textbf{p}_{i},  \hat{\textbf{e}}, \textbf{h}_{r}) = \hat{\textbf{e}} \cdot {\textbf{h}_{r}} \cdot \textbf{p}_{n} + (1-\hat{\textbf{e}} \cdot {\textbf{h}_{r}}) \cdot \textbf{p}_{i}
\end{equation}

where $\textbf{p}_{r} \in {[0,1]}^{B \times C}$ is the final outcome, with $B$ the batch size and $C$ the number of classes. The terms $\textbf{p}_{n}$ and $\textbf{p}_{i}$ are the output instances provided by the non-emotional and emotionally-influenced models, respectively, as already defined. The regulation function $R$ is learnable, selecting the evaluated outputs through a criterion, formally $\textbf{h}_{r} \in [0,1]^{B \times E}$, with $B$ the batch size and $E$ the emotional cardinality. The criterion is a matrix in which non-zero values represent the emotions to regulate. In particular, the matrix represents the expansion of a weight vector $\textbf{w}_{r} \in [0,1]^{1 \times E}$, in which elements define the regulation intensity associated with the considered emotions; formally, $\textbf{h}_r = Expand_{B,1}(\textbf{w}_{r})$, where $Expand$ is a function that replicates the input tensor $B$ times along the first dimension. Finally, the term $\hat{\textbf{e}} \in \{0,1\}^{B \times E}$ is a matrix of one-hot vectors in which non-null values correspond to maximum softmax probabilities in $\textbf{e}$.

Training is divided into three phases, each associated with a loss function.

\subsubsection{Emotional pre-training}
\label{emotional_pretraining}

In this phase, the model acquires artificial emotion by learning an affective history through the optimization of the emotional model. Formally,

\begin{equation}
        \mathcal{L}_{emot\_pre-train} = \mathcal{L}_{e}
    \end{equation}

where $\mathcal{L}_{e}$ is the cross-entropy loss related to the emotional outcome $\textbf{e}$. Optimization involves the weights of the Emotional Encoder and the related FCNN layers.

\subsubsection{Pre-training}
\label{pretraining}

The model learns the downstream classification task by optimizing non-emotional and emotionally-influenced models. Formally,

    \begin{equation}
        \mathcal{L}_{pre-train} \in \{\mathcal{L}_{n}, \mathcal{L}_{i}\}
    \end{equation}

where $\mathcal{L}_{n}$, and $\mathcal{L}_{i}$ are the cross-entropy loss functions related to predictions $\textbf{p}_{n}$, and $\textbf{p}_{i}$. The pre-training loss can assume the form of both the non-emotional and emotionally-influenced losses, optimized independently. Optimization involves the weights of the Non-emotional and Emotionally-influenced Encoders, together with the related fully-connected layers. 

\subsubsection{Training} 
\label{training}

The final phase optimizes classification by learning the Emotional Regulation process. The model learns the downstream classification task by optimizing non-emotional and emotionally-influenced models. Formally,

    \begin{equation}
        \mathcal{L}_{train} = \phi_{r} \mathcal{L}_{r} + \phi_{n} \mathcal{L}_{n} + \phi_{i} \mathcal{L}_{i}
    \end{equation}
    
where $\mathcal{L}_{r}$, $\mathcal{L}_{n}$, and $\mathcal{L}_{i}$ are the cross-entropy losses related to predictions $\textbf{p}_{r}$, $\textbf{p}_{n}$, and $\textbf{p}_{i}$, with weights $\phi_{r},\phi_{n}, \phi_{i} \in [0,1]$. Optimization involves the weights of the Non-emotional and Emotionally-influenced Encoders, together with the related fully-connected layers, as well as the parameters of the Emotional Regulation process. Weights in the loss function are considered as hyperparameters.

The emotional pre-training is considered mandatory, while the pre-training and training phases can be employed exclusively. In particular, Emotional Regulation can be performed without minimizing $\mathcal{L}_{train}$, since scores $\textbf{w}_{r}$ can be specified as hyperparameters. It is also possible to learn the regulation process without minimizing $\mathcal{L}_{pre-train}$, as non-emotional and emotionally-influenced models can be directly optimized during the training process involving $\mathcal{L}_{train}$. Some of the above training configurations are experimented through an ablation study, as described in the subsequent Sections.

\section{Experimental setup}
\label{experimental_setup}

\subsection{Architectures and learning configurations}
\label{models}

Experiments were conducted by employing two of the most widely employed architectures in deep learning, i.e., ResNet-50 \cite{he2016} and ViT-B/16 \cite{dosovitskiy2020}. These neural networks are suitable for evaluating the proposed approach with convolutional and transformer-based backbones, characterized by 25M and 86M parameters, respectively. Additional fully-connected layers were employed for mapping feature vectors linearly to softmax representations. Encoders pre-trained on ImageNet \cite{deng2009} were employed for all the experiments. 

Different learning configurations were trained and evaluated on CIFAR-10 and -100 datasets \cite{krizhevsky2009} for comparison with the original backbones. These represent two of the most widely employed databases for image classification benchmarking. The regulation approaches employed in the present study are described as follows.

\subsubsection{Non-learnable Emotional Regulation (E-Reg)} 
\label{non_learnable_emotional_regulation}

Pre-training is performed for the Non-emotional and Emotionally-influenced Encoders, while the Emotional Regulation process is non-learnable and characterized by constant regulation intensity $\textbf{w}_{r}$. These weights are defined by non-zero values associated with the dimensions related to emotions. Experiments involving the different combinations of the above emotional macro-classes were conducted. The approach does not involve the training phase, as described in the definition of the framework (Emotional Regulation).

\subsubsection{Learnable Emotional Regulation (Learn-E-Reg)}
\label{learnable_emotional_regulation}

Pre-training is performed for the Non-emotional and Emotionally-influenced Encoders. Emotional Regulation is learned during the training phase by optimizing $\textbf{w}_{r}$, initialized as a vector in which all the elements are equal to $0.5$, representing a \textit{half-regulation}.

\subsubsection{Learnable Emotional Regulation with Random Initialization (Rand-Learn-E-Reg)}
\label{randomly_learnable_emotional_regulation}

Pre-training is performed for the Non-emotional and Emotionally-influenced Encoders. Emotional Regulation is learned during training by optimizing $\textbf{w}_{r}$, initialized randomly at the beginning of each epoch.

\subsubsection{Fully Learnable Emotional Regulation (Full-Learn-E-Reg)}
\label{fully_learnable_emotional_regulation}

Pre-training is not performed. The Non-emotional and Emotionally-influenced Encoders are learned during the training phase. Emotional Regulation is learned during training as well, optimizing $\textbf{w}_{r}$, initialized as a vector in which all the elements are equal to $0.5$ (i.e., half-regulation).

\subsubsection{Fully Learnable Emotional Regulation with Random Initialization (Full-Rand-Learn-E-Reg)}
\label{fully_randomly_learnable_emotional_regulation}

Pre-training is not performed. The Non-emotional and Emotionally-influenced Encoders are learned during the training phase. Emotional Regulation is optimized during training as well, initializing $\textbf{w}_{r}$ randomly at the beginning of each epoch.

Emotional pre-training was performed for all the conducted experiments and chosen architectures.

\subsection{Datasets}
\label{datasets}

Several studies proposed architectures and frameworks for conducting classification of emotional stimuli by training deep learning models on image datasets \cite{devi2023, yang2021, zhang2025, liang2024, luo2025}. Image classification can be interpreted as an emotional learning process, assuming a labeled dataset as the set of emotional experiences. In the present study, four emotional datasets were employed to introduce an emotional history in the learning process. These sets of stimuli are described by different content and definitions of emotion. This variability allows for stronger generalization of the results throughout architectures and configurations of artificial subjective experience.

\subsubsection{EMOd}
\label{emod}

The EMOd (Emotional Attention Dataset) \cite{fan2018} is a set of emotion-evoking stimuli collected for studying the association between image emotion and visual attention. In the present work, 556 training and 143 validation samples of the dataset were considered, each associated with categorial classes describing the six universal emotions, as in the Ekman characterization \cite{ekman1992argument, ekman1971}, plus the neutral state. Emotions of \textit{anger}, \textit{fear}, \textit{sadness}, and \textit{disgust} were related to negative valence, while \textit{happiness} was associated with a positive outcome \cite{neta2023surprise}. The emotion of \textit{surprise} was associated with neutral valence, as in the case of the neutral state. In fact, valence related to surprise is debated \cite{neta2023surprise, speed2024ratings}; therefore, the present study assumes its semantics as neutral in order to avoid biases or ambiguity, as well as to limit the complexity of classes. 

A visualization of the EMOd dataset is provided in Fig. \ref{fig:combined_emotion}(b), where each sample was processed by BLIP \cite{li2022} for extracting language semantics, subsequently encoded through CLIP \cite{radford2021}, for image-text alignment, described by t-SNE \cite{maaten2008} components. Each instance of the dataset is associated with the related positive, neutral, and negative valence.

\subsubsection{Diffused-EMOd}
\label{diffused_emod}

Additional experiments involving an expansion of the EMOd dataset were conducted. In particular, the original training set was extended to 11,056 samples (i.e., 1,500 additional instances for each class) by training a diffusion model for synthetic image generation. A pre-trained architecture \cite{diffusion2022} (Stable Diffusion 1.4) was fine-tuned on the considered emotional stimuli by coupling each image in the dataset with the related emotional outcome. The resulting dataset considered the generated data as the training set, while the original EMOd validation set was reserved for testing. Experiments involving Diffused-EMOd allowed to investigate the generalization of the proposed approach with a higher quantity of stimuli, drawn from a distribution close to that of EMOd. Diffusion was learned by imposing 512x512 image resolution and batch size of 4, employing center cropping and random flipping as an augmentation strategy. Optimization was performed by employing AdamW \cite{adamw2017} with learning rate of $10^{-4}$. 

\subsubsection{Abstract}
\label{abstract}

A collection of abstract paintings associated with the emotional classes of \textit{amusement}, \textit{anger}, \textit{awe}, \textit{contentment}, \textit{disgust}, \textit{excitement}, \textit{fear}, and \textit{sadness} \cite{abstract2010}. The dataset is composed of 280 images characterized by different shapes and topics. Images were evaluated by human participants to vote for the emotional category. In the present work, 220 instances were associated with the training, while 60 were associated with the validation set by balancing occurrences among classes. Coherently with the original assumptions through which the dataset was defined \cite{abstract2010}, categories of amusement, awe, contentment, and excitement were considered as positive in valence, while anger, disgust, fear, and sadness were considered negative.

\subsubsection{EmoSet}
\label{emoset}

Large-scale balanced dataset for supporting the prediction of human emotional responses to visual stimuli \cite{emoset2023}. Images are associated with the classes of \textit{amusement}, \textit{awe}, \textit{contentment}, and \textit{excitement} for the positive valence, while \textit{anger}, \textit{disgust}, \textit{fear}, and \textit{sadness} for the negative. In the present study, a set of 95,480 samples was reserved for the training, while another set of 22,622 instances was reserved for validation. Also in this case, all the instances were balanced with respect to the split ratio.

\subsubsection{CIFAR-10 and -100}
\label{cifar_10_and_100}

Experiments were performed by training and evaluating downstream classification on CIFAR benchmarking datasets \cite{krizhevsky2009}. These two datasets are widely employed for image classification in deep learning, providing a suitable reference for the present study.

\subsection{Optimization and hyperparameter settings}
\label{experiments}

Non-emotional, emotionally-influenced, and emotional models were trained for 200 epochs with batch size of 32. Images were resized to 224x224 and augmented with random rotation up to 90 degrees; vertical and horizontal flipping was imposed with probability of 0.5. All the samples were normalized before being processed by the input layers. SGD (Stochastic Gradient Descent) \cite{sgd2016} was chosen as the optimizer for all the experiments, with learning rate of $10^{-3}$ and momentum of $0.9$. Weights related to $\mathcal{L}_{train}$ loss function were all defined as unity. Non-learnable regulation was evaluated by specifying regulation intensity as a hyperparameter, assuming different configurations associated with the positive, neutral, and negative semantics. For instance, the regulation of positive emotions was conducted by defining the related weights in $\textbf{w}_{r}$ as unity. All the experiments were conducted through NVIDIA Tesla V100 GPU instances from the CRESCO/ENEAGRID HPC cluster \cite{khan2023advanced, gebreyesus2024ai, chinnici2024towards}.

\section{Results}
\label{results}

In the experimentation, emotion was modeled as a concatenation of the original backbone embedding with features extracted from an emotional model. Emotional pre-training was conducted by employing four different datasets, including first-person real-life stimuli (EMOd \cite{fan2018}, EmoSet \cite{emoset2023}), abstract paintings (Abstract \cite{abstract2010}), and synthetic instances (Diffused-EMOd). This approach allows for introducing subjective emotional reactions in the learning process, overcoming the definition of emotion-augmented architectures strictly focused on neurophysiological factors. For comparing the results, downstream classification tasks were based on CIFAR-10 and -100 benchmarking datasets \cite{krizhevsky2009}. Different learning configurations were evaluated, involving random or pre-defined initialization of regulation intensity, as well as learnable or non-learnable Emotional Regulation and encoders. In particular, the \textit{EI} configuration involves the emotionally-influenced model only, while \textit{E-Reg} employs non-learnable Emotional Regulation. \textit{Learn-E-Reg} and \textit{Rand-Learn-E-Reg} assume frozen pre-trained encoders and learnable Emotional Regulation, while \textit{Full-Learn-E-Reg} and \textit{Full-Rand-Learn-E-Reg} realize end-to-end learning.

\subsection{Regulation based on EMOd}
\label{regulation_with_emod}

Emotional pre-training achieved 64.34\% and 65.03\% accuracy on EMOd by fine-tuning ResNet and ViT, respectively. The results show that improvements were found for all the dataset-architecture configurations. Emotional Regulation provided superior results with respect to the original backbones, proving the effectiveness of the proposed approach for improving image classification. Table \ref{results_cifar_emod} shows the accuracy results on CIFAR-10 and -100. Scores in bold highlight the best performance associated with the related configuration of dataset and architecture; underlined scores indicate performances overcoming the original backbone. 

\begin{table*}
\setcounter{table}{0}
\centering
\caption{Results on CIFAR-10 and -100 based on EMOd emotional history.}\label{results_cifar_emod}
\renewcommand{\arraystretch}{1.2} 
\setlength{\tabcolsep}{4pt} 
\resizebox{\textwidth}{!}{ 
\begin{tabular}{lllcccccc}
\toprule
\multicolumn{9}{c}{Accuracy on CIFAR-10 with Emotional Regulation based on EMOd emotional history} \\
\midrule
Model & Emot. Pre-train. Loss & Pre-train. Loss & Regulation & Regulation Loss & Positive & Neutral & Negative & Total \\
\midrule
\textit{ResNet-50 (backbone)} & None & $\mathcal{L}_{n}$ & None & None & 0.9353 & 0.9436 & 0.9154 & 0.9311 \\
EI-ResNet-50 & $\mathcal{L}_{e}$ & $\mathcal{L}_{i}$ & None & None & 0.9417 & 0.9447 & 0.9117 & \hlc[lightgreen]{0.9329} \\
E-Reg-ResNet-50 & $\mathcal{L}_{e}$ & $\mathcal{L}_{n}$, $\mathcal{L}_{i}$ & Neutral & None & 0.9417 & 0.9436 & 0.9117 & \hlc[lightgreen]{0.9326} \\
E-Reg-ResNet-50 & $\mathcal{L}_{e}$ & $\mathcal{L}_{n}$, $\mathcal{L}_{i}$ & Positive & None & 0.9353 & 0.9447 & 0.9117 & \hlc[lightred]{0.9302} \\
E-Reg-ResNet-50 & $\mathcal{L}_{e}$ & $\mathcal{L}_{n}$, $\mathcal{L}_{i}$ & Negative & None & 0.9417 & 0.9447 & 0.9154 & \hlc[lightgreen]{0.9341} \\
E-Reg-ResNet-50 & $\mathcal{L}_{e}$ & $\mathcal{L}_{n}$, $\mathcal{L}_{i}$ & Neutral, Positive & None & 0.9353 & 0.9436 & 0.9117 & \hlc[lightred]{0.9299} \\
E-Reg-ResNet-50 & $\mathcal{L}_{e}$ & $\mathcal{L}_{n}$, $\mathcal{L}_{i}$ & Neutral, Negative & None & 0.9417 & 0.9436 & 0.9154 & \hlc[lightgreen]{0.9338} \\
E-Reg-ResNet-50 & $\mathcal{L}_{e}$ & $\mathcal{L}_{n}$, $\mathcal{L}_{i}$ & Positive, Negative & None & 0.9353 & 0.9447 & 0.9154 & \hlc[lightgreen]{0.9314} \\
Learn-E-Reg-ResNet-50 & $\mathcal{L}_{e}$ & $\mathcal{L}_{n}$, $\mathcal{L}_{i}$ & Learned & $\mathcal{L}_{r}$ & 0.9479 & 0.9513 & 0.9276 & \hlc[lightgreen]{0.9423} \\
Rand-Learn-E-Reg-ResNet-50 & $\mathcal{L}_{e}$ & $\mathcal{L}_{n}$, $\mathcal{L}_{i}$ & Learned & $\mathcal{L}_{r}$ & 0.9491 & 0.9528 & 0.9267 & \hlc[lightgreen]{\textbf{0.9429}}\\
Full-Learn-E-Reg-ResNet-50 & $\mathcal{L}_{e}$ & None & Learned & $\mathcal{L}_{r}+\mathcal{L}_{n}+\mathcal{L}_{i}$ & 0.9518 & 0.9427 & 0.9303 & \hlc[lightgreen]{0.9410} \\
Full-Rand-Learn-E-Reg-ResNet-50 & $\mathcal{L}_{e}$ & None & Learned & $\mathcal{L}_{r}+\mathcal{L}_{n}+\mathcal{L}_{i}$ & 0.9485 & 0.9474 & 0.9220 & \hlc[lightgreen]{0.9397} \\
\midrule
\textit{ViT-B/16 (backbone)} & None & $\mathcal{L}_{n}$ & None & None & 0.9684 & 0.9775 & 0.9686 & 0.9702 \\
EI-ViT-B/16 & $\mathcal{L}_{e}$ & $\mathcal{L}_{i}$ & None & None & 0.9643 & 0.9781 & 0.9681 & \hlc[lightred]{0.9683} \\
E-Reg-ViT-B/16 & $\mathcal{L}_{e}$ & $\mathcal{L}_{n}$, $\mathcal{L}_{i}$ & Neutral & None & 0.9643 & 0.9775 & 0.9681 & \hlc[lightred]{0.9682} \\
E-Reg-ViT-B/16 & $\mathcal{L}_{e}$ & $\mathcal{L}_{n}$, $\mathcal{L}_{i}$ & Positive & None & 0.9684 & 0.9781 & 0.9681 & \hlc[lightred]{0.9701} \\
E-Reg-ViT-B/16 & $\mathcal{L}_{e}$ & $\mathcal{L}_{n}$, $\mathcal{L}_{i}$ & Negative & None & 0.9643 & 0.9781 & 0.9686 & \hlc[lightred]{0.9685} \\
E-Reg-ViT-B/16 & $\mathcal{L}_{e}$ & $\mathcal{L}_{n}$, $\mathcal{L}_{i}$ & Neutral, Positive & None & 0.9684 & 0.9775 & 0.9681 & \hlc[lightred]{0.9700} \\
E-Reg-ViT-B/16 & $\mathcal{L}_{e}$ & $\mathcal{L}_{n}$, $\mathcal{L}_{i}$ & Neutral, Negative & None & 0.9643 & 0.9775 & 0.9686 & \hlc[lightred]{0.9684} \\
E-Reg-ViT-B/16 & $\mathcal{L}_{e}$ & $\mathcal{L}_{n}$, $\mathcal{L}_{i}$ & Positive, Negative & None & 0.9684 & 0.9781 & 0.9686 & \hlc[lightgreen]{0.9703} \\
Learn-E-Reg-ViT-B/16 & $\mathcal{L}_{e}$ & $\mathcal{L}_{n}$, $\mathcal{L}_{i}$ & Learned & $\mathcal{L}_{r}$ & 0.9717 & 0.9797 & 0.9720 & \hlc[lightgreen]{\textbf{0.9733}} \\
Rand-Learn-E-Reg-ViT-B/16 & $\mathcal{L}_{e}$ & $\mathcal{L}_{n}$, $\mathcal{L}_{i}$ & Learned & $\mathcal{L}_{r}$ & 0.9715 & 0.9802 & 0.9717 & \hlc[lightgreen]{0.9732} \\
Full-Learn-E-Reg-ViT-B/16 & $\mathcal{L}_{e}$ & None & Learned & $\mathcal{L}_{r}+\mathcal{L}_{n}+\mathcal{L}_{i}$ & 0.9714 & 0.9774 & 0.9683 & \hlc[lightgreen]{0.9717} \\
Full-Rand-Learn-E-Reg-ViT-B/16 & $\mathcal{L}_{e}$ & None & Learned & $\mathcal{L}_{r}+\mathcal{L}_{n}+\mathcal{L}_{i}$ & 0.9703 & 0.9796 & 0.9672 & \hlc[lightgreen]{0.9706} \\
\midrule
\multicolumn{9}{c}{Accuracy on CIFAR-100 with Emotional Regulation based on EMOd emotional history} \\
\midrule
Model & Emot. Pre-train. Loss & Pre-train. Loss & Regulation & Regulation Loss & Positive & Neutral & Negative & Total \\
\midrule
\textit{ResNet-50 (backbone)} & None & $\mathcal{L}_{n}$ & None & None & 0.799 & 0.7635 & 0.7596 & 0.7768 \\
EI-ResNet-50 & $\mathcal{L}_{e}$ & $\mathcal{L}_{i}$ & None & None & 0.8094 & 0.766 & 0.7606 & \hlc[lightgreen]{0.7821} \\
E-Reg-ResNet-50 & $\mathcal{L}_{e}$ & $\mathcal{L}_{n}$, $\mathcal{L}_{i}$ & Neutral & None & 0.8094 & 0.7635 & 0.7606 & \hlc[lightgreen]{0.7813} \\
E-Reg-ResNet-50 & $\mathcal{L}_{e}$ & $\mathcal{L}_{n}$, $\mathcal{L}_{i}$ & Positive & None & 0.799 & 0.766 & 0.7606 & \hlc[lightgreen]{0.7779} \\
E-Reg-ResNet-50 & $\mathcal{L}_{e}$ & $\mathcal{L}_{n}$, $\mathcal{L}_{i}$ & Negative & None & 0.8094 & 0.766 & 0.7596 & \hlc[lightgreen]{0.7818} \\
E-Reg-ResNet-50 & $\mathcal{L}_{e}$ & $\mathcal{L}_{n}$, $\mathcal{L}_{i}$ & Neutral, Positive & None & 0.799 & 0.7635 & 0.7606 & \hlc[lightgreen]{0.7771} \\
E-Reg-ResNet-50 & $\mathcal{L}_{e}$ & $\mathcal{L}_{n}$, $\mathcal{L}_{i}$ & Neutral, Negative & None & 0.8094 & 0.7635 & 0.7596 & \hlc[lightgreen]{0.781} \\
E-Reg-ResNet-50 & $\mathcal{L}_{e}$ & $\mathcal{L}_{n}$, $\mathcal{L}_{i}$ & Positive, Negative & None & 0.799 & 0.766 & 0.7596 & \hlc[lightgreen]{0.7776} \\
Learn-E-Reg-ResNet-50 & $\mathcal{L}_{e}$ & $\mathcal{L}_{n}$, $\mathcal{L}_{i}$ & Learned & $\mathcal{L}_{r}$ & 0.8276 & 0.7847 & 0.7829 & \hlc[lightgreen]{\textbf{0.8016}} \\
Rand-Learn-E-Reg-ResNet-50 & $\mathcal{L}_{e}$ & $\mathcal{L}_{n}$, $\mathcal{L}_{i}$ & Learned & $\mathcal{L}_{r}$ & 0.8264 & 0.7831 & 0.7805 & \hlc[lightgreen]{0.7999} \\
Full-Learn-E-Reg-ResNet-50 & $\mathcal{L}_{e}$ & None & Learned & $\mathcal{L}_{r}+\mathcal{L}_{n}+\mathcal{L}_{i}$ & 0.8616 & 0.7932 & 0.7722 & \hlc[lightgreen]{0.7947} \\
Full-Rand-Learn-E-Reg-ResNet-50 & $\mathcal{L}_{e}$ & None & Learned & $\mathcal{L}_{r}+\mathcal{L}_{n}+\mathcal{L}_{i}$ & 0.8423 & 0.7945 & 0.764 & \hlc[lightgreen]{0.7933} \\
\midrule
\textit{ViT-B/16 (backbone)} & None & $\mathcal{L}_{n}$ & None & None & 0.8785 & 0.8454 & 0.8308 & 0.8543 \\
EI-ViT-B/16 & $\mathcal{L}_{e}$ & $\mathcal{L}_{i}$ & None & None & 0.8779 & 0.85 & 0.8248 & \hlc[lightred]{0.8541} \\
E-Reg-ViT-B/16 & $\mathcal{L}_{e}$ & $\mathcal{L}_{n}$, $\mathcal{L}_{i}$ & Neutral & None & 0.8779 & 0.8454 & 0.8248 & \hlc[lightred]{0.8525} \\
E-Reg-ViT-B/16 & $\mathcal{L}_{e}$ & $\mathcal{L}_{n}$, $\mathcal{L}_{i}$ & Positive & None & 0.8785 & 0.85 & 0.8248 & 0.8543 \\
E-Reg-ViT-B/16 & $\mathcal{L}_{e}$ & $\mathcal{L}_{n}$, $\mathcal{L}_{i}$ & Negative & None & 0.8779 & 0.85 & 0.8308 & \hlc[lightgreen]{0.8557} \\
E-Reg-ViT-B/16 & $\mathcal{L}_{e}$ & $\mathcal{L}_{n}$, $\mathcal{L}_{i}$ & Neutral, Positive & None & 0.8785 & 0.8454 & 0.8248 & \hlc[lightred]{0.8527} \\
E-Reg-ViT-B/16 & $\mathcal{L}_{e}$ & $\mathcal{L}_{n}$, $\mathcal{L}_{i}$ & Neutral, Negative & None & 0.8779 & 0.8454 & 0.8308 & \hlc[lightred]{0.8541} \\
E-Reg-ViT-B/16 & $\mathcal{L}_{e}$ & $\mathcal{L}_{n}$, $\mathcal{L}_{i}$ & Positive, Negative & None & 0.8785 & 0.85 & 0.8308 & \hlc[lightgreen]{0.8559} \\
Learn-E-Reg-ViT-B/16 & $\mathcal{L}_{e}$ & $\mathcal{L}_{n}$, $\mathcal{L}_{i}$ & Learned & $\mathcal{L}_{r}$ & 0.8844 & 0.8543 & 0.8391 & \hlc[lightgreen]{0.8619} \\
Rand-Learn-E-Reg-ViT-B/16 & $\mathcal{L}_{e}$ & $\mathcal{L}_{n}$, $\mathcal{L}_{i}$ & Learned & $\mathcal{L}_{r}$ & 0.8841 & 0.8566 & 0.8383 & \hlc[lightgreen]{\textbf{0.8624}} \\
Full-Learn-E-Reg-ViT-B/16 & $\mathcal{L}_{e}$ & None & Learned & $\mathcal{L}_{r}+\mathcal{L}_{n}+\mathcal{L}_{i}$ & 0.8821 & 0.8683 & 0.8506 & \hlc[lightgreen]{0.8622} \\
Full-Rand-Learn-E-Reg-ViT-B/16 & $\mathcal{L}_{e}$ & None & Learned & $\mathcal{L}_{r}+\mathcal{L}_{n}+\mathcal{L}_{i}$ & 0.8671 & 0.8617 & 0.8497 & \hlc[lightgreen]{0.8582} \\
\bottomrule
\end{tabular}}
\end{table*}

Learnable regulation achieved the best results, with Learn-E-Reg and Rand-Learn-E-Reg providing the most significant improvements. Learning regulation by employing pre-trained non-emotional and emotionally-influenced models allowed to overcome the original backbone by 1.27\% and 0.32\% on CIFAR-10 with ResNet and ViT, while 3.2\% and 0.95\% on CIFAR-100 with the same architectures, respectively. Table \ref{learned_regulation_emod} shows the learned regulation intensity for all the configurations, highlighting the overall best score in bold. The most accurate models are based on ViT, with parameters very close to half-regulation for CIFAR-10, while larger in magnitude towards negative emotions for CIFAR-100.

\begin{table*}[ht]
\centering
\caption{Learned regulation intensity with EMOd emotional history.}
\label{learned_regulation_emod}
\renewcommand{\arraystretch}{1.2} 
\setlength{\tabcolsep}{4pt}      
\resizebox{0.9\textwidth}{!}{%
\begin{tabular}{llcccccccc}
\toprule
\multicolumn{2}{c}{} & \multicolumn{7}{c}{Regulation intensity ($\textbf{w}_{r}$) with EMOd emotional history} \\
\cmidrule(lr){3-9}
Model & Dataset & Anger & Disgust & Fear & Happiness & Neutral & Sadness & Surprise & Accuracy \\
\midrule
Learn-E-Reg-ResNet-50         & CIFAR-10  & 0.5    & 0.5    & 0.5    & 0.5    & 0.5    & 0.5    & 0.5    & 0.9423 \\
Learn-E-Reg-ViT-B/16          & CIFAR-10  & 0.5001 & 0.5    & 0.5003 & 0.5001 & 0.5008 & 0.5003 & 0.5005 & \textbf{0.9733} \\
Rand-Learn-E-Reg-ResNet-50    & CIFAR-10  & 0.6127 & 0.1599 & 0.7078 & 0.2938 & 0.3465 & 0.7019 & 0.3642 & 0.9429 \\
Rand-Learn-E-Reg-ViT-B/16     & CIFAR-10  & 0.5453 & 0.8766 & 0.4564 & 0.569  & 0.7228 & 0.5234 & 0.7438 & 0.9732 \\
Full-Learn-E-Reg-ResNet-50     & CIFAR-10  & 0.4776 & 0.4793 & 0.4767 & 0.4732 & 0.48   & 0.4775 & 0.4691 & 0.941  \\
Full-Learn-E-Reg-ViT-B/16      & CIFAR-10  & 0.5001 & 0.5    & 0.5003 & 0.5001 & 0.5008 & 0.5003 & 0.5005 & 0.9717 \\
Full-Rand-Learn-E-Reg-ResNet-50 & CIFAR-10  & 0.1179 & 0.8151 & 0.8593 & 0.6424 & 0.6719 & 0.3715 & 0.1018 & 0.9397 \\
Full-Rand-Learn-E-Reg-ViT-B/16  & CIFAR-10  & 0.6836 & 0.5994 & 0.4231 & 0.2419 & 0.577  & 0.6364 & 0.4398 & 0.9706 \\
\midrule
Learn-E-Reg-ResNet-50         & CIFAR-100 & 0.5    & 0.5    & 0.5    & 0.5    & 0.5    & 0.5    & 0.5    & 0.8016 \\
Learn-E-Reg-ViT-B/16          & CIFAR-100 & 0.5263 & 0.4998 & 0.5021 & 0.5015 & 0.5112 & 0.4991 & 0.517  & 0.8619 \\
Rand-Learn-E-Reg-ResNet-50    & CIFAR-100 & 0.7605 & 0.6773 & 0.7125 & 0.4931 & 0.5943 & 0.3737 & 0.3389 & 0.7999 \\
Rand-Learn-E-Reg-ViT-B/16     & CIFAR-100 & 0.8013 & 0.7957 & 0.5008 & 0.1    & 0.3154 & 0.5547 & 0.3747 & \textbf{0.8624} \\
Full-Learn-E-Reg-ResNet-50     & CIFAR-100 & 0.4253 & 0.4246 & 0.4343 & 0.424  & 0.44   & 0.4339 & 0.4343 & 0.7947 \\
Full-Learn-E-Reg-ViT-B/16      & CIFAR-100 & 0.5263 & 0.4998 & 0.5021 & 0.5015 & 0.5112 & 0.4991 & 0.517  & 0.8622 \\
Full-Rand-Learn-E-Reg-ResNet-50 & CIFAR-100 & 0.1841 & 0.8378 & 0.7162 & 0.261  & 0.5293 & 0.732  & 0.7086 & 0.7933 \\
Full-Rand-Learn-E-Reg-ViT-B/16  & CIFAR-100 & 0.565  & 0.4235 & 0.774  & 0.0478 & 0.4338 & 0.3783 & 0.8043 & 0.8582 \\
\bottomrule
\end{tabular}%
}
\end{table*}

Non-learnable regulation (E-Reg) improved the original backbone by imposing the highest intensity for negative emotions. In the case of ViT, improvements were found by regulating both positive and negative semantics.

\subsection{Regulation based on Diffused-EMOd}
\label{regulation_with_diffused_emod}

Emotional Regulation provided significant results also by employing Diffused-EMOd for the emotional pre-training. The emotional model was trained on synthetic stimuli and evaluated on the original EMOd validation set, achieving accuracy scores of 55.24\% and 60.84\% by employing ResNet and ViT, respectively. Regulation provided improvements through both architectures, as shown in Table \ref{results_cifar_diff_emod}.

\begin{table*}
\centering
\caption{Results on CIFAR-10 and -100 based on Diffused-EMOd emotional history.}
\label{results_cifar_diff_emod}
\renewcommand{\arraystretch}{1.2} 
\setlength{\tabcolsep}{4pt} 
\resizebox{\textwidth}{!}{ 
\begin{tabular}{lllcccccc}
\toprule
\multicolumn{9}{c}{Accuracy on CIFAR-10 with Emotional Regulation based on Diffused-EMOd emotional history} \\
\midrule
Model & Emot. Pre-train. Loss & Pre-train. Loss & Regulation & Regulation Loss & Positive & Neutral & Negative & Total \\
\midrule
\textit{ResNet-50 (backbone)} & None & $\mathcal{L}_{n}$ & None & None & 0.9345 & 0.9426 & 0.9178 & 0.9311 \\
EI-ResNet-50 & $\mathcal{L}_{e}$ & $\mathcal{L}_{i}$ & None & None & 0.9364 & 0.9422 & 0.9193 & \hlc[lightgreen]{0.9323} \\
E-Reg-ResNet-50 & $\mathcal{L}_{e}$ & $\mathcal{L}_{n}$, $\mathcal{L}_{i}$ & Neutral & None & 0.9364 & 0.9426 & 0.9193 & \hlc[lightgreen]{0.9324} \\
E-Reg-ResNet-50 & $\mathcal{L}_{e}$ & $\mathcal{L}_{n}$, $\mathcal{L}_{i}$ & Positive & None & 0.9345 & 0.9422 & 0.9193 & \hlc[lightgreen]{0.9315} \\
E-Reg-ResNet-50 & $\mathcal{L}_{e}$ & $\mathcal{L}_{n}$, $\mathcal{L}_{i}$ & Negative & None & 0.9364 & 0.9422 & 0.9178 & \hlc[lightgreen]{0.9318} \\
E-Reg-ResNet-50 & $\mathcal{L}_{e}$ & $\mathcal{L}_{n}$, $\mathcal{L}_{i}$ & Neutral, Positive & None & 0.9345 & 0.9426 & 0.9193 & \hlc[lightgreen]{0.9316} \\
E-Reg-ResNet-50 & $\mathcal{L}_{e}$ & $\mathcal{L}_{n}$, $\mathcal{L}_{i}$ & Neutral, Negative & None & 0.9364 & 0.9426 & 0.9178 & \hlc[lightgreen]{0.9319} \\
E-Reg-ResNet-50 & $\mathcal{L}_{e}$ & $\mathcal{L}_{n}$, $\mathcal{L}_{i}$ & Positive, Negative & None & 0.9345 & 0.9422 & 0.9178 & \hlc[lightred]{0.931} \\
Learn-E-Reg-ResNet-50 & $\mathcal{L}_{e}$ & $\mathcal{L}_{n}$, $\mathcal{L}_{i}$ & Learned & $\mathcal{L}_{r}$ & 0.9471 & 0.953 & 0.9289 & \hlc[lightgreen]{0.9427} \\
Rand-Learn-E-Reg-ResNet-50 & $\mathcal{L}_{e}$ & $\mathcal{L}_{n}$, $\mathcal{L}_{i}$ & Learned & $\mathcal{L}_{r}$ & 0.9469 & 0.953 & 0.9293 & \hlc[lightgreen]{0.9427} \\
Full-Learn-E-Reg-ResNet-50 & $\mathcal{L}_{e}$ & None & Learned & $\mathcal{L}_{r}+\mathcal{L}_{n}+\mathcal{L}_{i}$ & 0.9473 & 0.9631 & 0.9372 & \hlc[lightgreen]{\textbf{0.9428}} \\
Full-Rand-Learn-E-Reg-ResNet-50 & $\mathcal{L}_{e}$ & None & Learned & $\mathcal{L}_{r}+\mathcal{L}_{n}+\mathcal{L}_{i}$ & 0.9406 & 0.9464 & 0.9369 & \hlc[lightgreen]{0.939} \\
\midrule
\textit{ViT-B/16 (backbone)} & None & $\mathcal{L}_{n}$ & None & None & 0.9705 & 0.9706 & 0.9691 & 0.9702 \\
EI-ViT-B/16 & $\mathcal{L}_{e}$ & $\mathcal{L}_{i}$ & None & None & 0.9713 & 0.9722 & 0.9622 & \hlc[lightred]{0.9693} \\
E-Reg-ViT-B/16 & $\mathcal{L}_{e}$ & $\mathcal{L}_{n}$, $\mathcal{L}_{i}$ & Neutral & None & 0.9713 & 0.9706 & 0.9622 & \hlc[lightred]{0.9689} \\
E-Reg-ViT-B/16 & $\mathcal{L}_{e}$ & $\mathcal{L}_{n}$, $\mathcal{L}_{i}$ & Positive & None & 0.9705 & 0.9722 & 0.9622 & \hlc[lightred]{0.9689} \\
E-Reg-ViT-B/16 & $\mathcal{L}_{e}$ & $\mathcal{L}_{n}$, $\mathcal{L}_{i}$ & Negative & None & 0.9713 & 0.9722 & 0.9691 & \hlc[lightgreen]{0.971} \\
E-Reg-ViT-B/16 & $\mathcal{L}_{e}$ & $\mathcal{L}_{n}$, $\mathcal{L}_{i}$ & Neutral, Positive & None & 0.9705 & 0.9706 & 0.9622 & \hlc[lightred]{0.9685} \\
E-Reg-ViT-B/16 & $\mathcal{L}_{e}$ & $\mathcal{L}_{n}$, $\mathcal{L}_{i}$ & Neutral, Negative & None & 0.9713 & 0.9706 & 0.9691 & \hlc[lightgreen]{0.9706} \\
E-Reg-ViT-B/16 & $\mathcal{L}_{e}$ & $\mathcal{L}_{n}$, $\mathcal{L}_{i}$ & Positive, Negative & None & 0.9705 & 0.9722 & 0.9691 & \hlc[lightgreen]{0.9706} \\
Learn-E-Reg-ViT-B/16 & $\mathcal{L}_{e}$ & $\mathcal{L}_{n}$, $\mathcal{L}_{i}$ & Learned & $\mathcal{L}_{r}$ & 0.9741 & 0.9754 & 0.9659 & \hlc[lightgreen]{0.9724} \\
Rand-Learn-E-Reg-ViT-B/16 & $\mathcal{L}_{e}$ & $\mathcal{L}_{n}$, $\mathcal{L}_{i}$ & Learned & $\mathcal{L}_{r}$ & 0.9737 & 0.9754 & 0.9687 & \hlc[lightgreen]{\textbf{0.9729}} \\
Full-Learn-E-Reg-ViT-B/16 & $\mathcal{L}_{e}$ & None & Learned & $\mathcal{L}_{r}+\mathcal{L}_{n}+\mathcal{L}_{i}$ & 0.9726 & 0.9796 & 0.9658 & \hlc[lightgreen]{0.9714} \\
Full-Rand-Learn-E-Reg-ViT-B/16 & $\mathcal{L}_{e}$ & None & Learned & $\mathcal{L}_{r}+\mathcal{L}_{n}+\mathcal{L}_{i}$ & 0.9742 & 0.965 & 0.9732 & \hlc[lightgreen]{0.9721} \\
\midrule
\multicolumn{9}{c}{Accuracy on CIFAR-100 with Emotional Regulation based on Diffused-EMOd emotional history} \\
\midrule
Model & Emot. Pre-train. Loss & Pre-train. Loss & Regulation & Regulation Loss & Positive & Neutral & Negative & Total \\
\midrule
\textit{ResNet-50 (backbone)} & None & $\mathcal{L}_{n}$ & None & None & 0.8056 & 0.7615 & 0.7544 & 0.7768 \\
EI-ResNet-50 & $\mathcal{L}_{e}$ & $\mathcal{L}_{i}$ & None & None & 0.8004 & 0.7721 & 0.7654 & \hlc[lightgreen]{0.7812} \\
E-Reg-ResNet-50 & $\mathcal{L}_{e}$ & $\mathcal{L}_{n}$, $\mathcal{L}_{i}$ & Neutral & None & 0.8004 & 0.7615 & 0.7654 & \hlc[lightgreen]{0.7783} \\
E-Reg-ResNet-50 & $\mathcal{L}_{e}$ & $\mathcal{L}_{n}$, $\mathcal{L}_{i}$ & Positive & None & 0.8056 & 0.7721 & 0.7654 & \hlc[lightgreen]{0.7833} \\
E-Reg-ResNet-50 & $\mathcal{L}_{e}$ & $\mathcal{L}_{n}$, $\mathcal{L}_{i}$ & Negative & None & 0.8004 & 0.7721 & 0.7544 & \hlc[lightgreen]{0.7776} \\
E-Reg-ResNet-50 & $\mathcal{L}_{e}$ & $\mathcal{L}_{n}$, $\mathcal{L}_{i}$ & Neutral, Positive & None & 0.8056 & 0.7615 & 0.7654 & \hlc[lightgreen]{0.7804} \\
E-Reg-ResNet-50 & $\mathcal{L}_{e}$ & $\mathcal{L}_{n}$, $\mathcal{L}_{i}$ & Neutral, Negative & None & 0.8004 & 0.7615 & 0.7544 & \hlc[lightred]{0.7747} \\
E-Reg-ResNet-50 & $\mathcal{L}_{e}$ & $\mathcal{L}_{n}$, $\mathcal{L}_{i}$ & Positive, Negative & None & 0.8056 & 0.7721 & 0.7544 & \hlc[lightgreen]{0.7797} \\
Learn-E-Reg-ResNet-50 & $\mathcal{L}_{e}$ & $\mathcal{L}_{n}$, $\mathcal{L}_{i}$ & Learned & $\mathcal{L}_{r}$ & 0.8254 & 0.7897 & 0.7813 & \hlc[lightgreen]{\textbf{0.8012}} \\
Rand-Learn-E-Reg-ResNet-50 & $\mathcal{L}_{e}$ & $\mathcal{L}_{n}$, $\mathcal{L}_{i}$ & Learned & $\mathcal{L}_{r}$ & 0.8226 & 0.7897 & 0.7819 & \hlc[lightgreen]{0.8003} \\
Full-Learn-E-Reg-ResNet-50 & $\mathcal{L}_{e}$ & None & Learned & $\mathcal{L}_{r}+\mathcal{L}_{n}+\mathcal{L}_{i}$ & 0.8313 & 0.7845 & 0.7919 & \hlc[lightgreen]{0.7984} \\
Full-Rand-Learn-E-Reg-ResNet-50 & $\mathcal{L}_{e}$ & None & Learned & $\mathcal{L}_{r}+\mathcal{L}_{n}+\mathcal{L}_{i}$ & 0.8364 & 0.8021 & 0.7796 & \hlc[lightgreen]{0.7952} \\
\midrule
\textit{ViT-B/16 (backbone)} & None & $\mathcal{L}_{n}$ & None & None & 0.8761 & 0.8445 & 0.8264 & 0.8543 \\
EI-ViT-B/16 & $\mathcal{L}_{e}$ & $\mathcal{L}_{i}$ & None & None & 0.8771 & 0.8467 & 0.8191 & \hlc[lightred]{0.854} \\
E-Reg-ViT-B/16 & $\mathcal{L}_{e}$ & $\mathcal{L}_{n}$, $\mathcal{L}_{i}$ & Neutral & None & 0.8771 & 0.8445 & 0.8191 & \hlc[lightred]{0.8532} \\
E-Reg-ViT-B/16 & $\mathcal{L}_{e}$ & $\mathcal{L}_{n}$, $\mathcal{L}_{i}$ & Positive & None & 0.8761 & 0.8467 & 0.8191 & \hlc[lightred]{0.8536} \\
E-Reg-ViT-B/16 & $\mathcal{L}_{e}$ & $\mathcal{L}_{n}$, $\mathcal{L}_{i}$ & Negative & None & 0.8771 & 0.8467 & 0.8264 & \hlc[lightgreen]{0.8555} \\
E-Reg-ViT-B/16 & $\mathcal{L}_{e}$ & $\mathcal{L}_{n}$, $\mathcal{L}_{i}$ & Neutral, Positive & None & 0.8761 & 0.8445 & 0.8191 & \hlc[lightred]{0.8528} \\
E-Reg-ViT-B/16 & $\mathcal{L}_{e}$ & $\mathcal{L}_{n}$, $\mathcal{L}_{i}$ & Neutral, Negative & None & 0.8771 & 0.8445 & 0.8264 & \hlc[lightgreen]{0.8547} \\
E-Reg-ViT-B/16 & $\mathcal{L}_{e}$ & $\mathcal{L}_{n}$, $\mathcal{L}_{i}$ & Positive, Negative & None & 0.8761 & 0.8467 & 0.8264 & \hlc[lightgreen]{0.8551} \\
Learn-E-Reg-ViT-B/16 & $\mathcal{L}_{e}$ & $\mathcal{L}_{n}$, $\mathcal{L}_{i}$ & Learned & $\mathcal{L}_{r}$ & 0.8815 & 0.8532 & 0.8341 & \hlc[lightgreen]{0.8614} \\
Rand-Learn-E-Reg-ViT-B/16 & $\mathcal{L}_{e}$ & $\mathcal{L}_{n}$, $\mathcal{L}_{i}$ & Learned & $\mathcal{L}_{r}$ & 0.8836 & 0.8543 & 0.8332 & \hlc[lightgreen]{\textbf{0.8625}} \\
Full-Learn-E-Reg-ViT-B/16 & $\mathcal{L}_{e}$ & None & Learned & $\mathcal{L}_{r}+\mathcal{L}_{n}+\mathcal{L}_{i}$ & 0.8588 & 0.8328 & 0.8735 & \hlc[lightgreen]{0.8617} \\
Full-Rand-Learn-E-Reg-ViT-B/16 & $\mathcal{L}_{e}$ & None & Learned & $\mathcal{L}_{r}+\mathcal{L}_{n}+\mathcal{L}_{i}$ & 0.8479 & 0.8472 & 0.8746 & \hlc[lightgreen]{0.8616} \\
\bottomrule
\end{tabular}}
\end{table*}

An improvement of 1.26\% was found on CIFAR-10 by employing fully learnable regulation (Full-Learn-E-Reg) based on ResNet, while 3.14\% more accuracy on CIFAR-100 was achieved through Learn-E-Reg based on the same architecture. In the case of ViT, the highest accuracy was reached by freezing pre-trained non-emotional and emotionally-influenced models, learning regulation through random intensity initialization (Rand-Learn-E-Reg). The approach overcame the backbone by 0.28\% and 0.96\% more accuracy on CIFAR-10 and -100, respectively. 

Models achieving the highest scores employed ViT as architecture, favoring the learning approach based on the random initialization of the regulation intensity. As shown in Table \ref{learned_regulation_diffused_emod}, training on CIFAR-10 resulted in a significant regulation of semantically contrasting emotions, such as \textit{fear} and \textit{happiness}. In the case of datasets characterized by a larger number of classes, such as CIFAR-100, a strong enhancement of the emotions of \textit{disgust}, \textit{happiness}, and \textit{sadness} was found. Non-learnable regulation (E-Reg) provided improvements when regulating negative emotions exclusively.

\begin{table*}[ht]
\centering
\caption{Learned regulation intensity with Diffused-EMOd emotional history.}
\label{learned_regulation_diffused_emod}
\renewcommand{\arraystretch}{1.2} 
\setlength{\tabcolsep}{4pt}      
\resizebox{0.9\textwidth}{!}{%
\begin{tabular}{llcccccccc}
\toprule
\multicolumn{2}{c}{} & \multicolumn{7}{c}{Regulation intensity ($\textbf{w}_{r}$) with Diffused-EMOd emotional history} \\
\cmidrule(lr){3-9}
Model & Dataset & Anger & Disgust & Fear & Happiness & Neutral & Sadness & Surprise & Accuracy \\
\midrule
Learn-E-Reg-ResNet-50         & CIFAR-10  & 0.5    & 0.5    & 0.5    & 0.5    & 0.5    & 0.5    & 0.5    & 0.9427 \\
Learn-E-Reg-ViT-B/16          & CIFAR-10  & 0.5    & 0.5    & 0.5002 & 0.5001 & 0.5035 & 0.5    & 0.503  & 0.9724 \\
Rand-Learn-E-Reg-ResNet-50    & CIFAR-10  & 0.3541 & 0.989  & 0.6938 & 0.4492 & 0.4415 & 0.2816 & 0.5033 & 0.9427 \\
Rand-Learn-E-Reg-ViT-B/16     & CIFAR-10  & 0.5133 & 0.5774 & 0.8006 & 0.8581 & 0.3223 & 0.3399 & 0.6839 & \textbf{0.9729} \\
Full-Learn-E-Reg-ResNet-50     & CIFAR-10  & 0.4723 & 0.4713 & 0.4565 & 0.464  & 0.4791 & 0.4556 & 0.4801 & 0.9428 \\
Full-Learn-E-Reg-ViT-B/16      & CIFAR-10  & 0.521  & 0.4984 & 0.5384 & 0.5184 & 0.5394 & 0.5358 & 0.5323 & 0.9714 \\
Full-Rand-Learn-E-Reg-ResNet-50 & CIFAR-10  & 0.5928 & 0.0339 & 0.2651 & 0.8538 & 0.437  & 0.394  & 0.3137 & 0.9390 \\
Full-Rand-Learn-E-Reg-ViT-B/16  & CIFAR-10  & 0.2991 & 0.0177 & 0.2491 & 0.7382 & 0.8254 & 0.598  & 0.3604 & 0.9721 \\
\midrule
Learn-E-Reg-ResNet-50         & CIFAR-100 & 0.5    & 0.5    & 0.5    & 0.5    & 0.5    & 0.5    & 0.5    & 0.8012 \\
Learn-E-Reg-ViT-B/16          & CIFAR-100 & 0.5002 & 0.5    & 0.4986 & 0.4998 & 0.4875 & 0.5003 & 0.4938 & 0.8614 \\
Rand-Learn-E-Reg-ResNet-50    & CIFAR-100 & 0.6599 & 0.2597 & 0.5639 & 0.129  & 0.5025 & 0.205  & 0.568  & 0.8003 \\
Rand-Learn-E-Reg-ViT-B/16     & CIFAR-100 & 0.5818 & 0.0826 & 0.2903 & 0.0388 & 0.575  & 0.0229 & 0.3891 & \textbf{0.8625} \\
Full-Learn-E-Reg-ResNet-50     & CIFAR-100 & 0.4344 & 0.4309 & 0.4339 & 0.4316 & 0.4526 & 0.4364 & 0.4427 & 0.7984 \\
Full-Learn-E-Reg-ViT-B/16      & CIFAR-100 & 0.5202 & 0.5021 & 0.5209 & 0.5323 & 0.537  & 0.5358 & 0.5068 & 0.8617 \\
Full-Rand-Learn-E-Reg-ResNet-50 & CIFAR-100 & 0.3158 & 0.7053 & 0.7841 & 0.3239 & 0.5779 & 0.5238 & 0.4663 & 0.7952 \\
Full-Rand-Learn-E-Reg-ViT-B/16  & CIFAR-100 & 0.5454 & 0.6176 & 0.3831 & 0.3877 & 0.4341 & 0.1531 & 0.7038 & 0.8616 \\
\bottomrule
\end{tabular}%
}
\end{table*}

\subsection{Regulation based on Abstract}
\label{regulation_with_abstract}

Emotional pre-training on Abstract achieved 45\% and 43.33\% accuracy through ResNet and ViT, respectively. Also in this experimentation, Emotional Regulation overcame the original backbone, as shown in Table \ref{results_cifar_abstract}. Regulation learning based on frozen ResNet encoders (Learn-E-Reg) improved by 1.37\% on CIFAR-10, while fully learnable regulation achieved 3.23\% more accuracy on CIFAR-100 by imposing random initialization (Full-Learn-E-Reg). In the case of ViT, frozen encoders and random intensity initialization (Rand-Learn-E-Reg) provided improvements of 0.34\% and 1.02\% on CIFAR-10 and -100, respectively. The non-learnable E-Reg approach provided improvements with respect to the backbone, except for the case of ViT on CIFAR-10.

\begin{table*}
\centering
\caption{Results on CIFAR-10 and -100 based on Abstract emotional history.}
\label{results_cifar_abstract}
\renewcommand{\arraystretch}{1.2} 
\setlength{\tabcolsep}{4pt} 
\resizebox{\textwidth}{!}{ 
\begin{tabular}{lllccccc}
\toprule
\multicolumn{8}{c}{Accuracy on CIFAR-10 with Emotional Regulation based on Abstract emotional history} \\
\midrule
Model & Emot. Pre-train. Loss & Pre-train. Loss & Regulation & Regulation Loss & Positive & Negative & Total \\
\midrule
\textit{ResNet-50 (backbone)} & None & $\mathcal{L}_{n}$ & None & None & 0.9367 & 0.9277 & 0.9311 \\
EI-ResNet-50 & $\mathcal{L}_{e}$ & $\mathcal{L}_{i}$ & None & None & 0.9431 & 0.9279 & \hlc[lightgreen]{0.9336} \\
E-Reg-ResNet-50 & $\mathcal{L}_{e}$ & $\mathcal{L}_{n}$, $\mathcal{L}_{i}$ & Positive & None & 0.9367 & 0.9279 & \hlc[lightgreen]{0.9312} \\
E-Reg-ResNet-50 & $\mathcal{L}_{e}$ & $\mathcal{L}_{n}$, $\mathcal{L}_{i}$ & Negative & None & 0.9431 & 0.9277 & \hlc[lightgreen]{0.9335} \\
Learn-E-Reg-ResNet-50 & $\mathcal{L}_{e}$ & $\mathcal{L}_{n}$, $\mathcal{L}_{i}$ & Learned & $\mathcal{L}_{r}$ & 0.9509 & 0.9397 & \hlc[lightgreen]{\textbf{0.9439}} \\
Rand-Learn-E-Reg-ResNet-50 & $\mathcal{L}_{e}$ & $\mathcal{L}_{n}$, $\mathcal{L}_{i}$ & Learned & $\mathcal{L}_{r}$ & 0.9509 & 0.9394 &\hlc[lightgreen]{0.9437} \\
Full-Learn-E-Reg-ResNet-50 & $\mathcal{L}_{e}$ & None & Learned & $\mathcal{L}_{r}+\mathcal{L}_{n}+\mathcal{L}_{i}$ & 0.9467 & 0.93 & \hlc[lightgreen]{0.9426} \\
Full-Rand-Learn-E-Reg-ResNet-50 & $\mathcal{L}_{e}$ & None & Learned & $\mathcal{L}_{r}+\mathcal{L}_{n}+\mathcal{L}_{i}$ & 0.94 & 0.9458 & \hlc[lightgreen]{0.9419} \\
\midrule
\textit{ViT-B/16 (backbone)} & None & $\mathcal{L}_{n}$ & None & None & 0.9713 & 0.9698 & 0.9702 \\
EI-ViT-B/16 & $\mathcal{L}_{e}$ & $\mathcal{L}_{i}$ & None & None & 0.968 & 0.9696 & \hlc[lightred]{0.9692} \\
E-Reg-ViT-B/16 & $\mathcal{L}_{e}$ & $\mathcal{L}_{n}$, $\mathcal{L}_{i}$ & Positive & None & 0.968 & 0.9698 & \hlc[lightred]{0.9693} \\
E-Reg-ViT-B/16 & $\mathcal{L}_{e}$ & $\mathcal{L}_{n}$, $\mathcal{L}_{i}$ & Negative & None & 0.9713 & 0.9696 & \hlc[lightred]{0.9701} \\
Learn-E-Reg-ViT-B/16 & $\mathcal{L}_{e}$ & $\mathcal{L}_{n}$, $\mathcal{L}_{i}$ & Learned & $\mathcal{L}_{r}$ & 0.9743 & 0.9725 & \hlc[lightgreen]{0.973} \\
Rand-Learn-E-Reg-ViT-B/16 & $\mathcal{L}_{e}$ & $\mathcal{L}_{n}$, $\mathcal{L}_{i}$ & Learned & $\mathcal{L}_{r}$ & 0.9747 & 0.9731 & \hlc[lightgreen]{\textbf{0.9735}} \\
Full-Learn-E-Reg-ViT-B/16 & $\mathcal{L}_{e}$ & None & Learned & $\mathcal{L}_{r}+\mathcal{L}_{n}+\mathcal{L}_{i}$ & 0.9672 & 0.9754 & \hlc[lightgreen]{0.9725} \\
Full-Rand-Learn-E-Reg-ViT-B/16 & $\mathcal{L}_{e}$ & None & Learned & $\mathcal{L}_{r}+\mathcal{L}_{n}+\mathcal{L}_{i}$ & 0.9692 & 0.9726 & \hlc[lightgreen]{0.9712} \\
\midrule
\multicolumn{8}{c}{Accuracy on CIFAR-100 with Emotional Regulation based on Abstract emotional history} \\
\midrule
Model & Emot. Pre-train. Loss & Pre-train. Loss & Regulation & Regulation Loss & Positive & Negative & Total \\
\midrule
\textit{ResNet-50 (backbone)} & None & $\mathcal{L}_{n}$ & None & None & 0.8072 & 0.7495 & 0.7768 \\
EI-ResNet-50 & $\mathcal{L}_{e}$ & $\mathcal{L}_{i}$ & None & None & 0.8122 & 0.7487 & \hlc[lightgreen]{0.7788} \\
E-Reg-ResNet-50 & $\mathcal{L}_{e}$ & $\mathcal{L}_{n}$, $\mathcal{L}_{i}$ & Positive & None & 0.8072 & 0.7487 & \hlc[lightred]{0.7764} \\
E-Reg-ResNet-50 & $\mathcal{L}_{e}$ & $\mathcal{L}_{n}$, $\mathcal{L}_{i}$ & Negative & None & 0.8122 & 0.7495 & \hlc[lightgreen]{0.7792} \\
Learn-E-Reg-ResNet-50 & $\mathcal{L}_{e}$ & $\mathcal{L}_{n}$, $\mathcal{L}_{i}$ & Learned & $\mathcal{L}_{r}$ & 0.8281 & 0.7721 & \hlc[lightgreen]{0.7986} \\
Rand-Learn-E-Reg-ResNet-50 & $\mathcal{L}_{e}$ & $\mathcal{L}_{n}$, $\mathcal{L}_{i}$ & Learned & $\mathcal{L}_{r}$ & 0.8285 & 0.7734 & \hlc[lightgreen]{0.7995} \\
Full-Learn-E-Reg-ResNet-50 & $\mathcal{L}_{e}$ & None & Learned & $\mathcal{L}_{r}+\mathcal{L}_{n}+\mathcal{L}_{i}$ & 0.8063 & 0.7908 & \hlc[lightgreen]{\textbf{0.8019}} \\
Full-Rand-Learn-E-Reg-ResNet-50 & $\mathcal{L}_{e}$ & None & Learned & $\mathcal{L}_{r}+\mathcal{L}_{n}+\mathcal{L}_{i}$ & 0.8013 & 0.7837 & \hlc[lightgreen]{0.7961} \\
\midrule
\textit{ViT-B/16 (backbone)} & None & $\mathcal{L}_{n}$ & None & None & 0.8642 & 0.8489 & 0.8543 \\
EI-ViT-B/16 & $\mathcal{L}_{e}$ & $\mathcal{L}_{i}$ & None & None & 0.8642 & 0.8506 & \hlc[lightgreen]{0.8554} \\
E-Reg-ViT-B/16 & $\mathcal{L}_{e}$ & $\mathcal{L}_{n}$, $\mathcal{L}_{i}$ & Positive & None & 0.8642 & 0.8506 & \hlc[lightgreen]{0.8554} \\
E-Reg-ViT-B/16 & $\mathcal{L}_{e}$ & $\mathcal{L}_{n}$, $\mathcal{L}_{i}$ & Negative & None & 0.8642 & 0.8489 & 0.8543 \\
Learn-E-Reg-ViT-B/16 & $\mathcal{L}_{e}$ & $\mathcal{L}_{n}$, $\mathcal{L}_{i}$ & Learned & $\mathcal{L}_{r}$ & 0.8699 & 0.8576 & \hlc[lightgreen]{0.8619} \\
Rand-Learn-E-Reg-ViT-B/16 & $\mathcal{L}_{e}$ & $\mathcal{L}_{n}$, $\mathcal{L}_{i}$ & Learned & $\mathcal{L}_{r}$ & 0.8713 & 0.8585 & \hlc[lightgreen]{\textbf{0.863}} \\
Full-Learn-E-Reg-ViT-B/16 & $\mathcal{L}_{e}$ & None & Learned & $\mathcal{L}_{r}+\mathcal{L}_{n}+\mathcal{L}_{i}$ & 0.8654 & 0.8585 & \hlc[lightgreen]{0.8614} \\
Full-Rand-Learn-E-Reg-ViT-B/16 & $\mathcal{L}_{e}$ & None & Learned & $\mathcal{L}_{r}+\mathcal{L}_{n}+\mathcal{L}_{i}$ & 0.8587 & 0.8594 & \hlc[lightgreen]{0.8594} \\
\bottomrule
\end{tabular}}
\end{table*}

The most accurate models involve random intensity initialization and frozen ViT encoders for learning Emotional Regulation. As shown in Table \ref{learned_regulation_abstract}, the configuration significantly regulated the emotions of \textit{amusement}, \textit{anger}, and \textit{awe}, while enhancing \textit{fear}, for CIFAR-10. Significant regulation was found for \textit{awe}, \textit{contentment}, and \textit{excitement} in the case of CIFAR-100, instead.

\begin{table*}[ht]
\centering
\caption{Learned regulation intensity with Abstract emotional history.}
\label{learned_regulation_abstract}
\renewcommand{\arraystretch}{1.2} 
\setlength{\tabcolsep}{4pt}      
\resizebox{\textwidth}{!}{%
\begin{tabular}{llccccccccc}
\toprule
\multicolumn{2}{c}{} & \multicolumn{8}{c}{Regulation intensity ($\textbf{w}_{r}$) with Abstract emotional history} \\
\cmidrule(lr){3-10}
Model & Dataset & Amusement & Anger & Awe & Contentment & Disgust & Excitement & Fear & Sadness & Accuracy \\
\midrule
Learn-E-Reg-ResNet-50         & CIFAR-10  & 0.5    & 0.5    & 0.5    & 0.5    & 0.5    & 0.5    & 0.5    & 0.5    & 0.9439 \\
Learn-E-Reg-ViT-B/16          & CIFAR-10  & 0.5    & 0.5    & 0.5    & 0.5034 & 0.5077 & 0.4999 & 0.5067 & 0.5079 & 0.9730 \\
Rand-Learn-E-Reg-ResNet-50    & CIFAR-10  & 0.5478 & 0.4317 & 0.6805 & 0.2819 & 0.1866 & 0.4186 & 0.5874 & 0.4828 & 0.9437 \\
Rand-Learn-E-Reg-ViT-B/16     & CIFAR-10  & 0.74   & 0.954  & 0.9876 & 0.656  & 0.3199 & 0.4376 & 0.0628 & 0.4343 & \textbf{0.9735} \\
Full-Learn-E-Reg-ResNet-50     & CIFAR-10  & 0.452  & 0.4926 & 0.4783 & 0.4674 & 0.4746 & 0.4728 & 0.4662 & 0.4598 & 0.9426 \\
Full-Learn-E-Reg-ViT-B/16      & CIFAR-10  & 0.5037 & 0.4996 & 0.5129 & 0.5189 & 0.5162 & 0.5354 & 0.5212 & 0.5276 & 0.9725 \\
Full-Rand-Learn-E-Reg-ResNet-50 & CIFAR-10  & 0.6848 & 0.3788 & 0.6501 & 0.4165 & 0.2097 & 0.4246 & 0.8751 & 0.1942 & 0.9419 \\
Full-Rand-Learn-E-Reg-ViT-B/16  & CIFAR-10  & 0.5826 & 0.4664 & 0.6417 & 0.6394 & 0.6692 & 0.6016 & 0.4847 & 0.4977 & 0.9712 \\
\midrule
Learn-E-Reg-ResNet-50         & CIFAR-100 & 0.5    & 0.5    & 0.5    & 0.5    & 0.5    & 0.5    & 0.5    & 0.5    & 0.7986 \\
Learn-E-Reg-ViT-B/16          & CIFAR-100 & 0.5    & 0.5    & 0.5001 & 0.5    & 0.4997 & 0.5    & 0.4999 & 0.5    & 0.8619 \\
Rand-Learn-E-Reg-ResNet-50    & CIFAR-100 & 0.6876 & 0.5624 & 0.4442 & 0.4308 & 0.5242 & 0.5415 & 0.4424 & 0.7016 & 0.7995 \\
Rand-Learn-E-Reg-ViT-B/16     & CIFAR-100 & 0.1934 & 0.6431 & 0.905  & 0.7249 & 0.4805 & 0.9308 & 0.2707 & 0.2639 & \textbf{0.863} \\
Full-Learn-E-Reg-ResNet-50     & CIFAR-100 & 0.4374 & 0.4942 & 0.4361 & 0.4448 & 0.4389 & 0.4466 & 0.4487 & 0.4482 & 0.8019 \\
Full-Learn-E-Reg-ViT-B/16      & CIFAR-100 & 0.6322 & 0.5084 & 0.7178 & 0.6369 & 0.6409 & 0.6373 & 0.6369 & 0.6496 & 0.8614 \\
Full-Rand-Learn-E-Reg-ResNet-50 & CIFAR-100 & 0.7484 & 0.338  & 0.2326 & 0.2145 & 0.7765 & 0.2453 & 0.5576 & 0.065  & 0.7961 \\
Full-Rand-Learn-E-Reg-ViT-B/16  & CIFAR-100 & 0.8029 & 0.9655 & 0.395  & 0.1904 & 0.7716 & 0.6151 & 0.3167 & 0.8176 & 0.8594 \\
\bottomrule
\end{tabular}%
}
\end{table*}

\subsection{Regulation based on EmoSet}
\label{regulation_with_emoset}

Experiments involving EmoSet revealed an emotional pre-training reaching 72.31\% and 75.47\% accuracy through ResNet and ViT, respectively. Results shown in Table \ref{results_cifar_emoset} reveal improvements with respect to backbones also in this case. Learning Emotional Regulation through frozen ResNet encoders provided improvements for both the target datasets. Random intensity initialization (Rand-Learn-E-Reg) achieved 1.57\% and 2.83\% more accuracy on CIFAR-10 and -100, respectively. The same configuration improved the ViT backbone by 0.30\% on CIFAR-10, while its version based on half-regulation initialization (Learn-E-Reg) achieved 0.91\% more accuracy on CIFAR-100. Non-learnable regulation (E-Reg) did not provide improvements in the case of ViT on CIFAR-100.

\begin{table*}
\centering
\caption{Results on CIFAR-10 and -100 based on EmoSet emotional history.}
\label{results_cifar_emoset}
\renewcommand{\arraystretch}{1.2} 
\setlength{\tabcolsep}{4pt} 
\resizebox{\textwidth}{!}{ 
\begin{tabular}{lllccccc}
\toprule
\multicolumn{8}{c}{Accuracy on CIFAR-10 with Emotional Regulation based on EmoSet emotional history} \\
\midrule
Model & Emot. Pre-train. Loss & Pre-train. Loss & Regulation & Regulation Loss & Positive & Negative & Total \\
\midrule
\textit{ResNet-50 (backbone)} & None & $\mathcal{L}_{n}$ & None & None & 0.9503 & 0.9304 & 0.9311 \\
EI-ResNet-50 & $\mathcal{L}_{e}$ & $\mathcal{L}_{i}$ & None & None & 0.9591 & 0.9365 & \hlc[lightgreen]{0.9373} \\
E-Reg-ResNet-50 & $\mathcal{L}_{e}$ & $\mathcal{L}_{n}$, $\mathcal{L}_{i}$ & Positive & None & 0.9503 & 0.9365 & \hlc[lightgreen]{0.937} \\
E-Reg-ResNet-50 & $\mathcal{L}_{e}$ & $\mathcal{L}_{n}$, $\mathcal{L}_{i}$ & Negative & None & 0.9591 & 0.9304 & \hlc[lightgreen]{0.9314} \\
Learn-E-Reg-ResNet-50 & $\mathcal{L}_{e}$ & $\mathcal{L}_{n}$, $\mathcal{L}_{i}$ & Learned & $\mathcal{L}_{r}$ & 0.9678 & 0.9442 & \hlc[lightgreen]{0.945} \\
Rand-Learn-E-Reg-ResNet-50 & $\mathcal{L}_{e}$ & $\mathcal{L}_{n}$, $\mathcal{L}_{i}$ & Learned & $\mathcal{L}_{r}$ & 0.9649 & 0.945 & \hlc[lightgreen]{\textbf{0.9457}} \\
Full-Learn-E-Reg-ResNet-50 & $\mathcal{L}_{e}$ & None & Learned & $\mathcal{L}_{r}+\mathcal{L}_{n}+\mathcal{L}_{i}$ & 0.9435 & 0.9414 & \hlc[lightgreen]{0.9429} \\
Full-Rand-Learn-E-Reg-ResNet-50 & $\mathcal{L}_{e}$ & None & Learned & $\mathcal{L}_{r}+\mathcal{L}_{n}+\mathcal{L}_{i}$ & 0.9186 & 0.9492 & \hlc[lightgreen]{0.9397} \\
\midrule
\textit{ViT-B/16 (backbone)} & None & $\mathcal{L}_{n}$ & None & None & 0.9757 & 0.9668 & 0.9702 \\
EI-ViT-B/16 & $\mathcal{L}_{e}$ & $\mathcal{L}_{i}$ & None & None & 0.976 & 0.9652 & \hlc[lightred]{0.9693} \\
E-Reg-ViT-B/16 & $\mathcal{L}_{e}$ & $\mathcal{L}_{n}$, $\mathcal{L}_{i}$ & Positive & None & 0.9757 & 0.9652 & \hlc[lightred]{0.9692} \\
E-Reg-ViT-B/16 & $\mathcal{L}_{e}$ & $\mathcal{L}_{n}$, $\mathcal{L}_{i}$ & Negative & None & 0.976 & 0.9668 & \hlc[lightgreen]{0.9703} \\
Learn-E-Reg-ViT-B/16 & $\mathcal{L}_{e}$ & $\mathcal{L}_{n}$, $\mathcal{L}_{i}$ & Learned & $\mathcal{L}_{r}$ & 0.9794 & 0.9681 & \hlc[lightgreen]{0.9724} \\
Rand-Learn-E-Reg-ViT-B/16 & $\mathcal{L}_{e}$ & $\mathcal{L}_{n}$, $\mathcal{L}_{i}$ & Learned & $\mathcal{L}_{r}$ & 0.9794 & 0.9694 & \hlc[lightgreen]{\textbf{0.9732}} \\
Full-Learn-E-Reg-ViT-B/16 & $\mathcal{L}_{e}$ & None & Learned & $\mathcal{L}_{r}+\mathcal{L}_{n}+\mathcal{L}_{i}$ & 0.9722 & 0.9706 & \hlc[lightgreen]{0.9716} \\
Full-Rand-Learn-E-Reg-ViT-B/16 & $\mathcal{L}_{e}$ & None & Learned & $\mathcal{L}_{r}+\mathcal{L}_{n}+\mathcal{L}_{i}$ & 0.9747 & 0.9626 & \hlc[lightred]{0.9699} \\
\midrule
\multicolumn{8}{c}{Accuracy on CIFAR-100 with Emotional Regulation based on EmoSet emotional history} \\
\midrule
Model & Emot. Pre-train. Loss & Pre-train. Loss & Regulation & Regulation Loss & Positive & Negative & Total \\
\midrule
\textit{ResNet-50 (backbone)} & None & $\mathcal{L}_{n}$ & None & None & 0.8309 & 0.7707 & 0.7768 \\
EI-ResNet-50 & $\mathcal{L}_{e}$ & $\mathcal{L}_{i}$ & None & None & 0.8476 & 0.7718 & \hlc[lightgreen]{0.7795} \\
E-Reg-ResNet-50 & $\mathcal{L}_{e}$ & $\mathcal{L}_{n}$, $\mathcal{L}_{i}$ & Positive & None & 0.8309 & 0.7718 & \hlc[lightgreen]{0.7778} \\
E-Reg-ResNet-50 & $\mathcal{L}_{e}$ & $\mathcal{L}_{n}$, $\mathcal{L}_{i}$ & Negative & None & 0.8476 & 0.7707 & \hlc[lightgreen]{0.7785} \\
Learn-E-Reg-ResNet-50 & $\mathcal{L}_{e}$ & $\mathcal{L}_{n}$, $\mathcal{L}_{i}$ & Learned & $\mathcal{L}_{r}$ & 0.8594 & 0.7918 & \hlc[lightgreen]{0.7987} \\
Rand-Learn-E-Reg-ResNet-50 & $\mathcal{L}_{e}$ & $\mathcal{L}_{n}$, $\mathcal{L}_{i}$ & Learned & $\mathcal{L}_{r}$ & 0.8663 & 0.7912 & \hlc[lightgreen]{\textbf{0.7988}} \\
Full-Learn-E-Reg-ResNet-50 & $\mathcal{L}_{e}$ & None & Learned & $\mathcal{L}_{r}+\mathcal{L}_{n}+\mathcal{L}_{i}$ & 0.8058 & 0.786 & \hlc[lightgreen]{0.7982} \\
Full-Rand-Learn-E-Reg-ResNet-50 & $\mathcal{L}_{e}$ & None & Learned & $\mathcal{L}_{r}+\mathcal{L}_{n}+\mathcal{L}_{i}$ & 0.7986 & 0.7895 & \hlc[lightgreen]{0.7952} \\
\midrule
\textit{ViT-B/16 (backbone)} & None & $\mathcal{L}_{n}$ & None & None & 0.876 & 0.8429 & 0.8543 \\
EI-ViT-B/16 & $\mathcal{L}_{e}$ & $\mathcal{L}_{i}$ & None & None & 0.8751 & 0.8398 & \hlc[lightred]{0.852} \\
E-Reg-ViT-B/16 & $\mathcal{L}_{e}$ & $\mathcal{L}_{n}$, $\mathcal{L}_{i}$ & Positive & None & 0.876 & 0.8398 & \hlc[lightred]{0.8523} \\
E-Reg-ViT-B/16 & $\mathcal{L}_{e}$ & $\mathcal{L}_{n}$, $\mathcal{L}_{i}$ & Negative & None & 0.8751 & 0.8429 & \hlc[lightred]{0.854} \\
Learn-E-Reg-ViT-B/16 & $\mathcal{L}_{e}$ & $\mathcal{L}_{n}$, $\mathcal{L}_{i}$ & Learned & $\mathcal{L}_{r}$ & 0.885 & 0.8501 & \hlc[lightgreen]{\textbf{0.8621}} \\
Rand-Learn-E-Reg-ViT-B/16 & $\mathcal{L}_{e}$ & $\mathcal{L}_{n}$, $\mathcal{L}_{i}$ & Learned & $\mathcal{L}_{r}$ & 0.8835 & 0.8505 & \hlc[lightgreen]{0.8619} \\
Full-Learn-E-Reg-ViT-B/16 & $\mathcal{L}_{e}$ & None & Learned & $\mathcal{L}_{r}+\mathcal{L}_{n}+\mathcal{L}_{i}$ & 0.8686 & 0.8556 & \hlc[lightgreen]{0.86} \\
Full-Rand-Learn-E-Reg-ViT-B/16 & $\mathcal{L}_{e}$ & None & Learned & $\mathcal{L}_{r}+\mathcal{L}_{n}+\mathcal{L}_{i}$ & 0.8725 & 0.8532 & \hlc[lightgreen]{0.8603} \\
\bottomrule
\end{tabular}}
\end{table*}

Models based on the vision transformer architecture provided the highest accuracy on both datasets, as shown in Table \ref{learned_regulation_emoset}. In particular, the configuration involving random intensity initialization with frozen encoders (Rand-Learn-E-Reg) achieved the best score on CIFAR-10. The same model with half-regulation as starting intensity (Learn-E-Reg) achieved the highest improvement on CIFAR-100, instead. In the first case, regulation was significant for the whole emotional spectrum, except for the category of \textit{excitement}, which was enhanced. In the second case, half-regulation intensity was predominant for all the emotional classes.

\begin{table*}[t]
\centering
\caption{Learned regulation intensity with EmoSet emotional history.}
\label{learned_regulation_emoset}
\renewcommand{\arraystretch}{1.2} 
\setlength{\tabcolsep}{4pt}      
\resizebox{\textwidth}{!}{%
\begin{tabular}{llccccccccc}
\toprule
\multicolumn{2}{c}{} & \multicolumn{8}{c}{Regulation intensity ($\textbf{w}_{r}$) with EmoSet emotional history} \\
\cmidrule(lr){3-10}
Model & Dataset & Amusement & Anger & Awe & Contentment & Disgust & Excitement & Fear & Sadness & Accuracy \\
\midrule
Learn-E-Reg-ResNet-50         & CIFAR-10  & 0.5    & 0.5    & 0.5    & 0.5    & 0.5    & 0.5    & 0.5    & 0.5    & 0.945 \\
Learn-E-Reg-ViT-B/16          & CIFAR-10  & 0.4969 & 0.4942 & 0.4897 & 0.4964 & 0.4829 & 0.4999 & 0.4848 & 0.4878 & 0.9724 \\
Rand-Learn-E-Reg-ResNet-50    & CIFAR-10  & 0.4311 & 0.6139 & 0.5862 & 0.6609 & 0.4545 & 0.9266 & 0.5383 & 0.3076 & 0.9457 \\
Rand-Learn-E-Reg-ViT-B/16     & CIFAR-10  & 0.7713 & 0.7167 & 0.7685 & 0.6211 & 0.9442 & 0.0358 & 0.6938 & 0.8280 & \textbf{0.9732} \\
Full-Learn-E-Reg-ResNet-50     & CIFAR-10  & 0.4667 & 0.4481 & 0.4677 & 0.4856 & 0.4698 & 0.4875 & 0.4691 & 0.4807 & 0.9429 \\
Full-Learn-E-Reg-ViT-B/16      & CIFAR-10  & 0.5076 & 0.5013 & 0.4935 & 0.5210 & 0.4876 & 0.5056 & 0.5196 & 0.5075 & 0.9716 \\
Full-Rand-Learn-E-Reg-ResNet-50 & CIFAR-10  & 0.1980 & 0.9446 & 0.7812 & 0.5809 & 0.1950 & 0.3012 & 0.5526 & 0.4286 & 0.9397 \\
Full-Rand-Learn-E-Reg-ViT-B/16  & CIFAR-10  & 0.6656 & 0.1225 & 0.1271 & 0.5868 & 0.2123 & 0.8833 & 0.8034 & 0.7726 & 0.9699 \\
\midrule
Learn-E-Reg-ResNet-50         & CIFAR-100 & 0.5    & 0.5    & 0.5    & 0.5    & 0.5    & 0.5    & 0.5    & 0.5    & 0.7987 \\
Learn-E-Reg-ViT-B/16          & CIFAR-100 & 0.5077 & 0.4995 & 0.5325 & 0.5040 & 0.5039 & 0.5001 & 0.5032 & 0.5033 & \textbf{0.8621} \\
Rand-Learn-E-Reg-ResNet-50    & CIFAR-100 & 0.0201 & 0.0551 & 0.4796 & 0.7393 & 0.6698 & 0.4761 & 0.5241 & 0.2123 & 0.7988 \\
Rand-Learn-E-Reg-ViT-B/16     & CIFAR-100 & 0.5384 & 0.1308 & 0.7343 & 0.5837 & 0.6798 & 0.6084 & 0.5450 & 0.1971 & 0.8619 \\
Full-Learn-E-Reg-ResNet-50     & CIFAR-100 & 0.4393 & 0.4391 & 0.4433 & 0.4395 & 0.4199 & 0.4365 & 0.4403 & 0.4432 & 0.7982 \\
Full-Learn-E-Reg-ViT-B/16      & CIFAR-100 & 0.6336 & 0.6207 & 0.6214 & 0.6429 & 0.5901 & 0.6353 & 0.5772 & 0.6045 & 0.8600 \\
Full-Rand-Learn-E-Reg-ResNet-50 & CIFAR-100 & 0.4272 & 0.3888 & 0.5453 & 0.2788 & 0.5827 & 0.2504 & 0.3935 & 0.5034 & 0.7933 \\
Full-Rand-Learn-E-Reg-ViT-B/16  & CIFAR-100 & 0.7532 & 0.4905 & 0.8041 & 0.9194 & 0.4334 & 0.9810 & 0.7468 & 0.6076 & 0.8603 \\
\bottomrule
\end{tabular}%
}
\end{table*}

\section{Discussion}
\label{discussion}

\subsection{Both non-emotional and emotionally-influenced representations are relevant for generalization}

The regulation of the neutral state involves neutral-related outcomes producing emotionally-influenced representations characterized by affective content. While the neutral outcome defines an emotion-informed state, its regulated representation, as provided by the Non-emotional Encoder, represents a non-emotion-informed state, i.e., an emotion-agnostic outcome. This is particularly evident for the non-learnable regulation (E-Reg), where the regulation intensity $\textbf{w}_{r}$ is defined through binary values.

Interestingly, the intensity of the regulation depends on the downstream dataset and the emotional history. In fact, the distribution of the parameters varies significantly among the various experiments. Therefore, artificial emotions improve classification learning through the particular (i.e., specific) affective history. There is no fixed set of stimuli or emotions improving learning, but such growth in accuracy is due to the impact of the given emotional history. The generalization process is supported (i.e., strengthened) by the custom artificial affectivity. It is possible to consider such characteristics of emotional specificity as a form of “artificial subjectivity” introduced in the model.

From the results, it is clear that the emotionally-influenced model alone does not always provide improvements. Emotionality as a concatenation of features does not ensure an improvement in the context of image classification. In fact, the regulation process was necessary to evaluate the emotional outcome, modulating the emotionally-influenced cognition. Approaches neglecting Emotional Regulation can be conceived as “affective reactivity,” through which emotion is dominant. The Emotional Regulation layer, therefore, introduces an additional cognitive factor improving generalization.

\subsection{Learnable Emotional Regulation overcomes the backbones}

In the case of non-learnable regulation (E-Reg), no constant improvements were found among the different experimental configurations of datasets and learning architectures. On one hand, the above approach allows for obtaining regulation without performing additional epochs; on the other hand, the strategy does not guarantee improvements with respect to the original backbone.

In all experiments, the Learn-E-Reg, Rand-Learn-E-Reg, and Full-Learn-E-Reg approaches introduced improvements. For the Full-Rand-Learn-E-Reg strategy, the optimization did not overcome the backbone on CIFAR-10 when pre-training the ViT-based Emotional Encoder on EmoSet. This suggests that the random factor in regulation introduces potential instability in the learning phase. It is possible that even the Rand-Learn-E-Reg approach, which likely exhibits the aforementioned characteristics, does not guarantee convergence towards an improved solution. However, experiments reveal that the configuration provided improvements across all the considered emotional datasets, especially for the case of EmoSet, for which it achieved the highest accuracy. Therefore, random intensity initialization was found to be effective when Emotional Regulation is learnable and all the encoders are pre-trained and frozen.

\subsection{Comparison with existing emotion-augmented models}

A comparison with studies in the related work is shown in Table \ref{sota_comparison}, in which, as can be seen, most of the studies focus on downstream tasks other than image classification. The latest works propose BEL models, adapted with specific architectures and backpropagation approaches, addressing time series forecasting problems. Other approaches involve EmBP for optimizing regression and classification tasks. A direct comparison is only possible with the study conducted by Zare et al. \cite{zare2022}, who experimented the AEmNN model on the CIFAR-10 and -100 datasets. Specifically, the authors reported average scores of 54.76\% and 27.53\% accuracy on the validation set, respectively. Our approach overcomes the above performances, reaching 97.35\% and 86.3\% accuracy through the Rand-Learn-E-Reg-ViT-B/16 model, establishing Emotional Regulation as the new state-of-the-art in emotion-augmented deep learning on the aforementioned CIFAR datasets.

\begin{table*}[t]
\centering
\caption{Comparison of Emotional Regulation with the related work in emotion-augmented deep learning.}
\label{sota_comparison}
\renewcommand{\arraystretch}{1.2}
\setlength{\tabcolsep}{4pt}
\resizebox{\textwidth}{!}{%
\begin{tabular}{llllll}
\toprule
Model & Method & Emotion Model & Downstream Task & Dataset & Performance Metrics \\
\midrule
PSO-EmNN \raisebox{0pt}{\cite{shahid2020}} & 
\begin{tabular}[c]{@{}l@{}}Brain Emotional Learning (BEL);\\ 
Particle Swarm Optimization (PSO)\end{tabular} & 
\begin{tabular}[c]{@{}l@{}}Anatomical structure of the\\ 
emotional brain\end{tabular} & 
\begin{tabular}[c]{@{}l@{}}Coronary artery disease\\ 
diagnosis (classification)\end{tabular} & 
Z-Alizadeh Sani & 
\begin{tabular}[c]{@{}l@{}}Accuracy, Precision,\\ 
Sensitivity, Specificity,\\ 
F1-score\end{tabular} \\
\midrule
CD-ELN \raisebox{0pt}{\cite{xu2021}} & 
\begin{tabular}[c]{@{}l@{}}Brain Emotional Learning (BEL);\\ 
context-dependent modules;\\ 
memristive circuit implementation\end{tabular} & 
\begin{tabular}[c]{@{}l@{}}Anatomical structure of the\\ 
emotional brain\end{tabular} & 
Multitask classification & 
\begin{tabular}[c]{@{}l@{}}Custom dataset\\ 
(digit images);\\ 
MNIST\end{tabular} & 
Accuracy \\
\midrule
GA-EANN \raisebox{0pt}{\cite{abba2022}} & 
\begin{tabular}[c]{@{}l@{}}Emotional Backpropagation (EmBP);\\ 
genetic algorithm\end{tabular} & 
\begin{tabular}[c]{@{}l@{}}Hormonal parameters related\\ 
to anxiety and confidence states\end{tabular} & 
\begin{tabular}[c]{@{}l@{}}Water quality index\\ 
prediction\end{tabular} & 
\begin{tabular}[c]{@{}l@{}}Custom dataset\\ 
(Kinta River, Malaysia)\end{tabular} & 
\begin{tabular}[c]{@{}l@{}}MAE, MSE, RMSE,\\ 
MAPE, NSE, CC\end{tabular} \\
\midrule
AEmNN \raisebox{0pt}{\cite{zare2022}} & 
\begin{tabular}[c]{@{}l@{}}Emotional Backpropagation (EmBP);\\ 
dopamine-inspired adaptive learning;\\ 
stochastic learning\end{tabular} & 
\begin{tabular}[c]{@{}l@{}}Anxiety and confidence states\\ 
modulated by simulated\\ 
dopamine regulation\end{tabular} & 
Image classification & 
\begin{tabular}[c]{@{}l@{}}ORL, Yale, Yale-B, MIT,\\ 
MNIST, Fashion-MNIST,\\ 
CIFAR-10/100, SVHN,\\ 
CINIC-10\end{tabular} & 
\begin{tabular}[c]{@{}l@{}}Accuracy, Precision,\\ 
Recall, F1-score,\\
R-squared, MAE\end{tabular} \\
\midrule
B2ELM-BEL \raisebox{0pt}{\cite{suthasinee2024}} & 
\begin{tabular}[c]{@{}l@{}}Brain Emotional Learning (BEL);\\ 
Biased Extreme Learning Machines\\ 
(Biased-ELM); knowledge transfer\end{tabular} & 
\begin{tabular}[c]{@{}l@{}}Anatomical structure of the\\ 
emotional brain\end{tabular} & 
Time series forecasting & 
\begin{tabular}[c]{@{}l@{}}Chaotic time series\\ 
(Henon, Lorenz,\\ 
Rossler, etc.); real-world\\ 
energy consumption\\ 
datasets\end{tabular} & 
MAE, RMSE, SMAPE \\
\midrule
RWBELNN \raisebox{0pt}{\cite{kumar2005}} & 
\begin{tabular}[c]{@{}l@{}}Brain Emotional Learning (BEL);\\ 
recurrent weighted connections;\\ 
Lyapunov-stability-based adaptive\\ 
Backpropagation\end{tabular} & 
\begin{tabular}[c]{@{}l@{}}Anatomical structure of the\\ 
emotional brain\end{tabular} & 
Time series forecasting & 
\begin{tabular}[c]{@{}l@{}}Chaotic time series\\ 
(synthetic nonlinear plant)\end{tabular} & 
MAE, MSE, RMSE \\
\midrule
\begin{tabular}[c]{@{}l@{}}\textit{Emotional Regulation}\\ 
\textit{(ours)}\end{tabular} & 
\begin{tabular}[c]{@{}l@{}}Weighted sum of non-emotional \\
and emotionally-influenced responses\\ 
based on emotional outcome;\\ 
pre-training on emotional stimuli\end{tabular} & 
\begin{tabular}[c]{@{}l@{}}Subjective emotional\\ 
experience\end{tabular} & 
Image classification & 
CIFAR-10/100 & 
Accuracy \\
\bottomrule
\end{tabular}%
}
\end{table*}

As can be noticed, the related work focuses on neurophysiological aspects of emotion, investigating models inspired by hormonal parameters, dopamine regulation, and anatomical structures of the brain. In this context, our framework introduces a novel approach inspired by subjective experience for modeling artificial emotion in deep learning.

\subsection{Subjective experience as a significant factor for distinguishing human and machine emotion}

Beyond the results obtained in terms of improved classification performance, the present study suggests some important observations. In particular, it is relevant that emotion is characterized by subjective aspects \cite{bechara2004role, paul2020towards}. Indeed, it is difficult to establish a priori the emotion associated with a given visual stimulus. Although some sensory instances are more commonly associated with certain emotional reactions, other visual configurations are not immediately predictable in affective terms. For instance, consider a subject looking at a landscape featuring an active volcano. The emotional reaction of the subject above cannot be traced a priori to a specific perception. While a person may experience the emotion of fear due to the dangerousness of the volcano, another subject may experience a state closer to happiness, as the view of the landscape can be perceived as pleasant. In this context, while the Emotional Regulation framework requires a pre-established set of stimulus-emotion associations, in humans such a correspondence is not always predictable. Furthermore, the affective reaction can be dynamic in humans, as a stimulus can be perceived differently over time. The same effect was not found in the proposed Emotional Regulation model, for which a given stimulus always corresponds to a single, unchanging affective state.

The aforementioned characteristics indicate a significant difference between human and artificial emotion, as proposed in the present study. The investigation aims to provide hypotheses for future research. Further experiments may involve the modeling of architectures allowing for emotional pre-training through sets of non-predefined -- yet still consistent -- emotional stimuli. Furthermore, it would be interesting to define computational processes characterizing dynamic affective states over time, assuming variation criteria based on emotional history and the incremental evaluation of visual instances.

An interesting path for modeling emotion, as well as the related emotional history, involves the identification of a homeostatic process through which a deep neural network -- or an agent, in general -- can express autonomous reactions to stimuli. In this context, Man \& Damasio \cite{man2019homeostasis} proposed well-being as a criterion of the homeostatic process. We say that providing an agent with a structural criterion of well-being, defined, e.g., as the need to be itself or in an optimal state, may allow the elicitation of non-predefined emotional reactions. Such a homeostatic model, coupled with the Emotional Regulation mechanism, could enhance the learning process in a deep neural network. However, emotional classes -- or affective states, in general -- may be different from those defined for humans. Notice that although emotional stimuli are predetermined in the present study, the proposed framework exhibits characteristics and effects that resemble a homeostatic behavior. Specifically, the learning of regulation intensity parameters can be interpreted as the neural network’s attempt to self-achieve emotional balance, enabling more effective learning. The results of the present study show that a model entirely influenced by emotion (i.e., the emotionally-influenced model) does not always reach improvements in learning. To ensure better generalization, it is necessary to introduce a regulation mechanism.

The definition of a homeostatic model provides the opportunity to deepen the understanding of the fundamental criteria of acting and feeling, with the ultimate goal of self-realization. In the case that the fundamental criterion of homeostasis is represented by the need for well-being, how can this factor be defined in artificial agents? Furthermore, what definition can be given of well-being itself? A possible research horizon could investigate homeostasis as the need of a being to fulfil itself, be itself, or achieve a specific goal. In human beings, it seems that such a criterion is originally provided by nature; in machines, this factor may instead be defined by humans.

\subsection{Artificial subjective emotion improves deep learning-based image classification}

Results provide evidence that learnable regulation, expressed through the configurations of Learn-E-Reg, Rand-Learn-E-Reg, and Full-Learn-E-Reg, improves image classification in deep learning. In particular, the models reaching the highest accuracy score outperformed the related ViT backbone through an increase of 0.34\% and 1.02\% on CIFAR-10 and -100, respectively. On the same datasets, Emotional Regulation outperformed the ResNet backbone by achieving 1.57\% and 3.19\% higher accuracy. Furthermore, a comparison with the related work reveals that Emotional Regulation overcomes existing solutions by margins of 77.77\% and 213.48\% in accuracy on the above two benchmarks, respectively. In this context, the proposed framework represents the new state-of-the-art in emotion-augmented deep learning on large-scale vision datasets. Emotional Regulation represents a novel method for enhancing generalization in computer vision tasks, as well as a valuable approach for introducing artificial subjective emotion in learning systems. The study provides further evidence that emotion improves deep learning-based image classification, highlighting the importance of affective history and subjective experience in emotion-augmented models.

\subsection{Horizons for future work}

Further investigation could address the optimization of the emotional learning process. The present study considers emotional training in terms of categorical variables related to affective states. Future work may explore training the Emotional Encoder employing continuous label distributions \cite{zhang2025, liang2024}. Other studies may assume dimensional models of emotion, such as those characterized by valence and arousal \cite{lang1995emotion, russell1980circumplex}, as well as other significant variables \cite{mehrabian1997comparison}. Instead of predicting target classes, the Emotional Encoder could be trained on associations defined by continuous quantities, optimizing a regression task. The Emotional Encoder can be prone to biases due to data scarcity or imbalanced visual stimuli. A promising approach could involve the use of generative models to create synthetic datasets that are less prone to biases. In this context, the work conducted by Yang et al. \cite{yang2025} enables the generation of different emotional variations of the same image. This process can be employed to reduce unfairness in the emotional dataset.

A future study may also concern the investigation of Emotional Regulation by integrating state-of-the-art methods in the field of emotion-augmented deep learning. Specifically, encoders in the framework can be integrated with BEL architectures \cite{shahid2020, parvinizadeh2022, suthasinee2024} --  considering the anatomical structure of the emotional brain -- and emotional backpropagation \cite{zare2022} -- simulating processes related to dopamine, anxiety, and confidence. Studying human cognitive processes is fundamental for modeling the significant factors of emotion, as well as for identifying suitable applications to validate the proposed method. Along this path, Keating \& Cook \cite{keating2023} provided evidence for the dependence of emotional expression recognition on internal affective states; an investigation involving Emotional Regulation can be conducted to improve facial expression recognition in the wild. Furthermore, future studies may involve the introduction of additional emotional factors inspired by human cognitive processes, such as models defining saccadic eye movements \cite{zhang2025}.

Beyond serving as a framework for demonstrating the significant impact of emotion in deep learning, as well as an approach for improving generalization in vision tasks, the proposed model aims to enhance the definition of affectivity in artificial agents through emotional pre-training. This approach can be applied in the context of human-robot interaction (HRI), where the identification of emotional patterns is crucial for context interpretation and decision-making \cite{moerland2018emotion, newman2025empathy}. Emotional Regulation can be employed to optimize a vision task performed by an agent, as well as to provide semantic understanding of stimuli acquired from the environment. The dataset employed to train the Emotional Encoder would represent the knowledge through which a robot associates affective semantics with world instances, enhancing context adaptation and interaction with humans. The set of emotional stimuli could also serve as a potential appraisal criterion for modeling the empathic process. The elicited emotion can be modulated and enriched by information regarding the affective state of external subjects, with the aim of integrating Emotional Regulation with incremental knowledge. In addition to the emotion resulting from the processing of a visual stimulus, the emotionally-influenced representation could be conditioned on features related to facial expressions, visual and linguistic contexts, as well as other factors influencing empathy \cite{landi2020computational}.

The above perspective is also important for addressing ethical concerns related to emotionally responsive artificial intelligence systems \cite{BenZion2025WhyWN}. Emotional pre-training can be adapted to a set of stimuli describing behaviors of emotional dependency or signals of attachment in human-machine interaction. Aligning images and text representations enables the estimation of the emotional content of interaction, improving empathy, and defining guardrails based on perspective-taking and other relevant processes.

\section*{Acknowledgments}

This work is supported by NOVA LINCS (UID/04516/2025) with the financial support of FCT.IP, and by the project AI.INSIGHTS: AI-Powered Real-Time Anonymous User Echo Monitoring (ALGARVE-FEDER-02964500, Ref. 24298) co-financed by ALGARVE 2030, Portugal 2030 and by the European Union.

Marta Chinnici was supported for this research by Project ECS 0000024 Rome Technopole, -- CUP B83C22002820006, National Recovery and Resilience Plan (NRRP), Mission 4, Component 2 Investment 1.5", funded by the European Union -- NextGenerationEU. João Rodrigues was supported by UID/04516/NOVA Laboratory for Computer Science and Informatics (NOVA LINCS) with the financial support of Portuguese Foundation for Science and Technology (FCT.IP).

The computing resources and the related technical support used for this work have been provided by CRESCO/ENEAGRID High Performance Computing infrastructure and its staff. CRESCO/ENEAGRID High Performance Computing infrastructure is funded by ENEA, the Italian National Agency for New Technologies, Energy and Sustainable Economic Development and by Italian and European research programmes; see \href{http://www.cresco.enea.it/english}{http://www.cresco.enea.it/english} for information.

\renewcommand*{\bibfont}{\footnotesize}

\setlength{\biblabelsep}{0.5em}   
\setlength{\bibhang}{1.6em}       
\setlength{\labelnumberwidth}{1.6em} 

\defbibenvironment{bibliography}
  {\list
     {\printfield[labelnumberwidth]{labelnumber}}
     {%
      \setlength{\labelwidth}{\labelnumberwidth}%
      \setlength{\leftmargin}{\labelwidth}%
      \setlength{\labelsep}{0.5em}
      \addtolength{\leftmargin}{\labelsep}%
      \setlength{\itemsep}{0pt}%
      \setlength{\parsep}{0pt}%
      \setlength{\parskip}{0pt}%
     }%
  }
  {\endlist}
  {\item}

\setlength\bibinitsep{0.0em}
\printbibliography[title={References}]

@article{keating2023,
  title={The inside out model of emotion recognition: how the shape of one’s internal emotional landscape influences the recognition of others’ emotions},
  author={Keating, C. T. and Cook, J. L.},
  journal={Scientific Reports},
  volume={13},
  number={1},
  pages={21490},
  year={2023},
  publisher={Nature Publishing Group UK London}
}

@article{zadra2011,
  title={Emotion and perception: The role of affective information},
  author={Zadra, J. R. and Clore, G. L.},
  journal={Wiley Interdisciplinary Reviews: Cognitive Science},
  volume={2},
  number={6},
  pages={676--685},
  year={2011},
  publisher={Wiley Online Library}
}

@article{becker2011,
  title={Attentional selection is biased toward mood-congruent stimuli},
  author={Becker, M. W. and Leinenger, M.},
  journal={Emotion},
  volume={11},
  number={5},
  pages={1248},
  year={2011},
  publisher={American Psychological Association}
}

@article{vanhoff2011,
  title={Disgust- and not fear-evoking images hold our attention},
  author={Van Hooff, J. C. and Devue, C. and Vieweg, P. E. and Theeuwes, J.},
  journal={Acta Psychologica},
  volume={143},
  number={1},
  pages={1--6},
  year={2013},
  publisher={Elsevier}
}

@article{yin2023,
  title={The effects of emotion on judgments of learning and memory: A meta-analytic review},
  author={Yin, Y. and Shanks, D. R. and Li, B. and Fan, T. and Hu, X. and Yang, C. and Luo, L.},
  journal={Metacognition and Learning},
  volume={18},
  number={2},
  pages={425--447},
  year={2023},
  publisher={Springer}
}

@article{tyng2017,
  title={The influences of emotion on learning and memory},
  author={Tyng, C. M. and Amin, H. U. and Saad, M. N. M. and Malik, A. S.},
  journal={Frontiers in Psychology},
  volume={8},
  pages={235933},
  year={2017},
  publisher={Frontiers}
}

@article{congleton2020,
  title={The devil is in the details: Investigating the influence of emotion on event memory using a simulated event},
  author={Congleton, A. R. and Berntsen, D.},
  journal={Psychological Research},
  volume={84},
  number={8},
  pages={2339--2353},
  year={2020},
  publisher={Springer}
}

@article{brosch2013impact,
  title={The impact of emotion on perception, attention, memory, and decision-making},
  author={Brosch, T. and Scherer, K. and Grandjean, D. and Sander, D.},
  journal={Swiss Medical Weekly},
  volume={143},
  number={1920},
  pages={w13786--w13786},
  year={2013}
}

@inproceedings{stromfelt2017,
  title={Emotion-augmented machine learning: Overview of an emerging domain},
  author={Str{\"o}mfelt, H. and Zhang, Y. and Schuller, B. W.},
  booktitle={2017 Seventh International Conference on Affective Computing and Intelligent Interaction (ACII)},
  pages={305--312},
  year={2017},
  organization={IEEE}
}

@inproceedings{lotfi2012,
  title={Supervised brain emotional learning},
  author={Lotfi, E. and Akbarzadeh-T, M.-R.},
  booktitle={The 2012 International Joint Conference on Neural Networks (IJCNN)},
  pages={1--6},
  year={2012},
  organization={IEEE}
}

@article{lotfi2013,
  title={Brain emotional learning-based pattern recognizer},
  author={Lotfi, E. and Akbarzadeh-T, M.-R.},
  journal={Cybernetics and Systems},
  volume={44},
  number={5},
  pages={402--421},
  year={2013},
  publisher={Taylor \& Francis}
}

@article{parvinizadeh2022,
  title={A simple and efficient rainfall--runoff model based on supervised brain emotional learning},
  author={Parvinizadeh, S. and Zakermoshfegh, M. and Shakiba, M.},
  journal={Neural Computing and Applications},
  volume={34},
  number={2},
  pages={1509--1526},
  year={2022},
  publisher={Springer}
}

@article{kremer2019,
  title={Influence of negative emotions on residents’ learning of scientific information: an experimental study},
  author={Kremer, T. and Mamede, S. and van den Broek, W. W. and Schmidt, H. G. and Nunes, M. P. T. and Martins, M. A.},
  journal={Perspectives on Medical Education},
  volume={8},
  number={4},
  pages={209--215},
  year={2019},
  publisher={Springer}
}

@article{shahid2020,
  title={A novel approach for coronary artery disease diagnosis using hybrid particle swarm optimization based emotional neural network},
  author={Shahid, A. H. and Singh, M. P.},
  journal={Biocybernetics and Biomedical Engineering},
  volume={40},
  number={4},
  pages={1568--1585},
  year={2020},
  publisher={Elsevier}
}

@article{zamirpour2018,
  title={A biological brain-inspired fuzzy neural network: Fuzzy emotional neural network},
  author={Zamirpour, E. and Mosleh, M.},
  journal={Biologically Inspired Cognitive Architectures},
  volume={26},
  pages={80--90},
  year={2018},
  publisher={Elsevier}
}

@article{suthasinee2024,
  title={Bias-boosted ELM for knowledge transfer in brain emotional learning for time series forecasting},
  author={Iamsa-At, S. and Horata, P. and Sunat, K.},
  journal={IEEE Access},
  volume={12},
  pages={35868--35898},
  year={2024},
  publisher={IEEE}
}

@article{li2020role,
  title={The role of positive emotions in education: A neuroscience perspective},
  author={Li, L. and Gow, A. D. I. and Zhou, J.},
  journal={Mind, Brain and Education},
  volume={14},
  number={3},
  pages={220--234},
  year={2020},
  publisher={Wiley Online Library}
}

@article{chinnici2024towards,
  title={Towards Sustainability and Energy Efficiency Using Data Analytics for HPC Data Center},
  author={Chinnici, A. and Ahmadzada, E. and Kor, A.-L. and De Chiara, D. and Dom{\'\i}nguez-D{\'\i}az, A. and de Marcos Ortega, L. and Chinnici, M.},
  journal={Electronics},
  volume={13},
  number={17},
  pages={3542},
  year={2024},
  publisher={MDPI}
}

@article{zhou2025improved,
  title={Improved IEC performance via emotional stimuli-aware captioning},
  author={Zhou, Z. and Zhai, Z. and Gao, X. and Zhu, J.},
  journal={Scientific Reports},
  volume={15},
  number={1},
  pages={22173},
  year={2025},
  publisher={Nature Publishing Group UK London}
}

@article{zhou2025object,
  title={Object-scene semantics correlation analysis for image emotion classification},
  author={Zhou, Z. and Zhai, Z. and Chen, H. and Lu, S.},
  journal={Frontiers in Neuroscience},
  volume={19},
  pages={1657562},
  year={2025},
  publisher={Frontiers}
}

@article{gebreyesus2024ai,
  title={AI for automating data center operations: model explainability in the data centre context using shapley additive explanations (SHAP)},
  author={Gebreyesus, Y. and Dalton, D. and De Chiara, D. and Chinnici, M. and Chinnici, A.},
  journal={Electronics},
  volume={13},
  number={9},
  pages={1628},
  year={2024},
  publisher={MDPI}
}

@article{khan2023advanced,
  title={Advanced data analytics modeling for evidence-based data center energy management},
  author={Khan, W. and De Chiara, D. and Kor, A.-L. and Chinnici, M.},
  journal={Physica A: Statistical Mechanics and its Applications},
  volume={624},
  pages={128966},
  year={2023},
  publisher={Elsevier}
}

@article{kumar2005,
  title={New recurrent weighted lyapunov-stability based brain emotional learning-based neural network: Application to the modeling of the nonlinear dynamical system},
  author={Kumar, R.},
  journal={Circuits, Systems, and Signal Processing},
  volume={44},
  number={9},
  pages={6467--6493},
  year={2025},
  publisher={Springer}
}

@article{BenZion2025WhyWN,
  title={Why we need mandatory safeguards for emotionally responsive AI},
  author={Ben-Zion, Z.},
  journal={Nature},
  year={2025},
  volume={643},
  pages={9 - 9}
}

@article{pan2025positive,
  title={Positive Emotion Enhances Memory by Promoting Memory Reinstatement across Repeated Learning},
  author={Pan, R. and Gao, C. and Zhu, X. and Li, B. and Jia, X.},
  journal={Journal of Neuroscience},
  volume={45},
  number={31},
  year={2025},
  publisher={Society for Neuroscience}
}

@inproceedings{wang2025enhancing,
  title={Enhancing emotion reasoning for image multi-emotion prediction},
  author={Wang, B. and Tu, G. and Liang, B. and Bai, Z. and Yang, M. and Zeng, X. and Yao, L. and Xu, R.},
  booktitle={ICASSP 2025-2025 IEEE International Conference on Acoustics, Speech and Signal Processing (ICASSP)},
  pages={1--5},
  year={2025},
  organization={IEEE}
}

@article{xu2021,
  title={Memristive circuit implementation of context-dependent emotional learning network and its application in multitask},
  author={Xu, C. and Wang, C. and Jiang, J. and Sun, J. and Lin, H.},
  journal={IEEE Transactions on Computer-Aided Design of Integrated Circuits and Systems},
  volume={41},
  number={9},
  pages={3052--3065},
  year={2021},
  publisher={IEEE}
}

@article{embp2008,
  title={A modified backpropagation learning algorithm with added emotional coefficients},
  author={Khashman, A.},
  journal={IEEE Transactions on Neural Networks},
  volume={19},
  number={11},
  pages={1896--1909},
  year={2008},
  publisher={IEEE}
}

@article{embp2009,
  title={Application of an emotional neural network to facial recognition},
  author={Khashman, A.},
  journal={Neural Computing and Applications},
  volume={18},
  number={4},
  pages={309--320},
  year={2009},
  publisher={Springer}
}

@inproceedings{thenius2013,
  title={EMANN-a model of emotions in an artificial neural network},
  author={Thenius, R. and Zahadat, P. and Schmickl, T.},
  booktitle={Artificial Life Conference Proceedings},
  pages={830--837},
  year={2013},
  organization={MIT Press One Rogers Street, Cambridge, MA 02142-1209, USA}
}

@article{abba2022,
  title={Integrating feature extraction approaches with hybrid emotional neural networks for water quality index modeling},
  author={Abba, S. I. and Abdulkadir, R. A. and Sammen, S. S. and Pham, Q. B. and Lawan, A. A. and Esmaili, P. and Malik, A. and Al-Ansari, N.},
  journal={Applied Soft Computing},
  volume={114},
  pages={108036},
  year={2022},
  publisher={Elsevier}
}

@article{zare2022,
  title={A dopamine based adaptive emotional neural network},
  author={Zare, M. A. and Boostani, R. and Mohammadi, M. and Kouchaki, S.},
  journal={IEEE Access},
  volume={10},
  pages={109460--109475},
  year={2022},
  publisher={IEEE}
}

@article{devi2023,
  title={Deep convolutional neural networks with transfer learning for visual sentiment analysis},
  author={Usha Kingsly Devi, K. and Gomathi, V.},
  journal={Neural Processing Letters},
  volume={55},
  number={4},
  pages={5087--5120},
  year={2023},
  publisher={Springer}
}

@article{yang2021,
  title={SOLVER: Scene-object interrelated visual emotion reasoning network},
  author={Yang, J. and Gao, X. and Li, L. and Wang, X. and Ding, J.},
  journal={IEEE Transactions on Image Processing},
  volume={30},
  pages={8686--8701},
  year={2021},
  publisher={IEEE}
}

@article{zhang2025,
  title={Saccade inspired Attentive Visual Patch Transformer for image sentiment analysis},
  author={Zhang, J. and Zhu, J. and Sun, H. and Zhang, X. and Liu, J.},
  journal={Applied Soft Computing},
  volume={174},
  pages={112963},
  year={2025},
  publisher={Elsevier}
}

@article{liang2024,
  title={Non-uniform circular-structured loss inspired by psychology for image emotion recognition},
  author={Liang, Z. and Li, H. and Zhang, R. and Liu, X.},
  journal={Multimedia Systems},
  volume={30},
  number={6},
  pages={346},
  year={2024},
  publisher={Springer}
}

@article{luo2025,
  title={CVRSF-Net: Image emotion recognition by combining visual relationship features and scene features},
  author={Luo, Y. and Zhong, X. and Xie, J. and Liu, G.},
  journal={IEEE Transactions on Emerging Topics in Computational Intelligence},
  year={2025},
  publisher={IEEE}
}

@inproceedings{landi2024,
  title={An investigation of the impact of emotion in image classification based on deep learning},
  author={Landi, R. E. and Chinnici, M. and Iovane, G.},
  booktitle={International Conference on Human-Computer Interaction},
  pages={300--310},
  year={2024},
  organization={Springer}
}

@inproceedings{he2016,
  title={Deep residual learning for image recognition},
  author={He, K. and Zhang, X. and Ren, S. and Sun, J.},
  booktitle={Proceedings of the IEEE Conference on Computer Vision and Pattern Recognition},
  pages={770--778},
  year={2016}
}

@article{dosovitskiy2020,
  title={An image is worth 16x16 words: Transformers for image recognition at scale},
  author={Dosovitskiy, A. and Beyer, L. and Kolesnikov, A. and Weissenborn, D. and Zhai, X. and Unterthiner, T. and Dehghani, M. and Minderer, M. and Heigold, G. and Gelly, S. and others},
  journal={arXiv preprint arXiv:2010.11929},
  year={2020}
}

@inproceedings{deng2009,
  title={ImageNet: A large-scale hierarchical image database},
  author={Deng, J. and Dong, W. and Socher, R. and Li, L.-J. and Li, K. and Fei-Fei, L.},
  booktitle={2009 IEEE Conference on Computer Vision and Pattern Recognition},
  pages={248--255},
  year={2009},
  organization={Ieee}
}

@inproceedings{fan2018,
  title={Emotional attention: A study of image sentiment and visual attention},
  author={Fan, S. and Shen, Z. and Jiang, M. and Koenig, B. L. and Xu, J. and Kankanhalli, M. S. and Zhao, Q.},
  booktitle={Proceedings of the IEEE Conference on Computer Vision and Pattern Recognition},
  pages={7521--7531},
  year={2018}
}

@article{ekman1971,
  title={Constants across cultures in the face and emotion.},
  author={Ekman, P. and Friesen, W. V.},
  journal={Journal of Personality and Social Psychology},
  volume={17},
  number={2},
  pages={124},
  year={1971},
  publisher={American Psychological Association}
}

@inproceedings{li2022,
  title={BLIP: Bootstrapping language-image pre-training for unified vision-language understanding and generation},
  author={Li, J. and Li, D. and Xiong, C. and Hoi, S.},
  booktitle={International Conference on Machine Learning},
  pages={12888--12900},
  year={2022},
  organization={PMLR}
}

@inproceedings{radford2021,
  title={Learning transferable visual models from natural language supervision},
  author={Radford, A. and Kim, J. W. and Hallacy, C. and Ramesh, A. and Goh, G. and Agarwal, S. and Sastry, G. and Askell, A. and Mishkin, P. and Clark, J. and others},
  booktitle={International Conference on Machine Learning},
  pages={8748--8763},
  year={2021},
  organization={PmLR}
}

@article{maaten2008,
  title={Visualizing data using t-SNE},
  author={van der Maaten, L. and Hinton, G.},
  journal={Journal of Machine Learning Research},
  volume={9},
  number={Nov},
  pages={2579--2605},
  year={2008}
}

@article{diffusion2022,
  title={Photorealistic text-to-image diffusion models with deep language understanding},
  author={Saharia, C. and Chan, W. and Saxena, S. and Li, L. and Whang, J. and Denton, E. L. and Ghasemipour, K. and Gontijo Lopes, R. and Karagol Ayan, B. and Salimans, T. and others},
  journal={Advances in Neural Information Processing Systems},
  volume={35},
  pages={36479--36494},
  year={2022}
}

@article{adamw2017,
  title={Decoupled weight decay regularization},
  author={Loshchilov, I. and Hutter, F.},
  journal={arXiv preprint arXiv:1711.05101},
  year={2017}
}

@inproceedings{abstract2010,
  title={Affective image classification using features inspired by psychology and art theory},
  author={Machajdik, J. and Hanbury, A.},
  booktitle={Proceedings of the 18th ACM International Conference on Multimedia},
  pages={83--92},
  year={2010}
}

@inproceedings{emoset2023,
  title={EmoSet: A large-scale visual emotion dataset with rich attributes},
  author={Yang, J. and Huang, Q. and Ding, T. and Lischinski, D. and Cohen-Or, D. and Huang, H.},
  booktitle={Proceedings of the IEEE/CVF International Conference on Computer Vision},
  pages={20383--20394},
  year={2023}
}

@inproceedings{man2020truth,
  title={Truth or consequences: Homeostatic self-regulation in artificial neural networks},
  author={Man, K. and Damasio, A.},
  booktitle={Artificial Life Conference Proceedings 32},
  pages={146--147},
  year={2020},
  organization={MIT Press One Rogers Street, Cambridge, MA 02142-1209, USA}
}

@article{landi2020computational,
  title={A computational model of cognitive empathy based on incremental learning and the analysis of facial micro-expressions and minimal gesture cues},
  author={Landi, R. E.},
  year={2020}
}

@article{man2019homeostasis,
  title={Homeostasis and soft robotics in the design of feeling machines},
  author={Man, K. and Damasio, A.},
  journal={Nature Machine Intelligence},
  volume={1},
  number={10},
  pages={446--452},
  year={2019},
  publisher={Nature Publishing Group UK London}
}

@article{lang1995emotion,
  title={The emotion probe: Studies of motivation and attention.},
  author={Lang, P. J.},
  journal={American Psychologist},
  volume={50},
  number={5},
  pages={372},
  year={1995},
  publisher={American Psychological Association}
}

@article{mehrabian1997comparison,
  title={Comparison of the PAD and PANAS as models for describing emotions and for differentiating anxiety from depression},
  author={Mehrabian, A.},
  journal={Journal of Psychopathology and Behavioral Assessment},
  volume={19},
  number={4},
  pages={331--357},
  year={1997},
  publisher={Springer}
}

@article{russell1980circumplex,
  title={A circumplex model of affect},
  author={Russell, J. A.},
  journal={Journal of Personality and Social Psychology},
  volume={39},
  number={6},
  pages={1161},
  year={1980},
  publisher={American Psychological Association}
}

@article{ekman1992argument,
  title={An argument for basic emotions},
  author={Ekman, Paul},
  journal={Cognition \& Emotion},
  volume={6},
  number={3-4},
  pages={169--200},
  year={1992},
  publisher={Taylor \& Francis}
}

@inproceedings{jacobs2014emergent,
  title={Emergent dynamics of joy, distress, hope and fear in reinforcement learning agents},
  author={Jacobs, E. and Broekens, J. and Jonker, C.},
  booktitle={Adaptive Learning Agents Workshop at AAMAS2014},
  year={2014}
}

@techreport{krizhevsky2009,
  title={Learning multiple layers of features from tiny images},
  author={Krizhevsky, A. and Hinton, G. and others},
  year={2009},
  publisher={Toronto, ON, Canada}
}

@article{paul2020towards,
  title={Towards a comparative science of emotion: Affect and consciousness in humans and animals},
  author={Paul, E. S. and Sher, S. and Tamietto, M. and Winkielman, P. and Mendl, M. T.},
  journal={Neuroscience \& Biobehavioral Reviews},
  volume={108},
  pages={749--770},
  year={2020},
  publisher={Elsevier}
}

@article{khashman2011credit,
  title={Credit risk evaluation using neural networks: Emotional versus conventional models},
  author={Khashman, A.},
  journal={Applied Soft Computing},
  volume={11},
  number={8},
  pages={5477--5484},
  year={2011},
  publisher={Elsevier}
}

@article{brown2020coherence,
  title={Coherence between subjective experience and physiology in emotion: Individual differences and implications for well-being},
  author={Brown, C. L. and Van Doren, N. and Ford, B. Q. and Mauss, I. B. and Sze, J. W. and Levenson, R. W.},
  journal={Emotion},
  volume={20},
  number={5},
  pages={818},
  year={2020},
  publisher={American Psychological Association}
}

@article{zhang2025neurofunctional,
  title={A neurofunctional signature of affective arousal generalizes across valence domains and distinguishes subjective experience from autonomic reactivity},
  author={Zhang, R. and Gan, X. and Xu, T. and Yu, F. and Wang, L. and Song, X. and Jiao, G. and Liu, X. and Zhou, F. and Becker, B.},
  journal={Nature Communications},
  volume={16},
  number={1},
  pages={6492},
  year={2025},
  publisher={Nature Publishing Group UK London}
}

@article{neta2023surprise,
  title={Surprise as an emotion: A response to Ortony},
  author={Neta, M. and Kim, M. J.},
  journal={Perspectives on Psychological Science},
  volume={18},
  number={4},
  pages={854--862},
  year={2023},
  publisher={Sage Publications Sage CA: Los Angeles, CA}
}

@article{speed2024ratings,
  title={Ratings of valence, arousal, happiness, anger, fear, sadness, disgust, and surprise for 24,000 Dutch words},
  author={Speed, L. J. and Brysbaert, M.},
  journal={Behavior Research Methods},
  volume={56},
  number={5},
  pages={5023--5039},
  year={2024},
  publisher={Springer}
}

@article{bechara2004role,
  title={The role of emotion in decision-making: Evidence from neurological patients with orbitofrontal damage},
  author={Bechara, A.},
  journal={Brain and Cognition},
  volume={55},
  number={1},
  pages={30--40},
  year={2004},
  publisher={Elsevier}
}

@article{newman2025empathy,
  title={Empathy in Long-Term Human--Robot Interaction: A Scoping Review of Emotion Understanding},
  author={Newman, M. S. and Senadji, B. and White, K. M. and Fookes, C.},
  journal={International Journal of Social Robotics},
  volume={17},
  number={1},
  pages={191--210},
  year={2025},
  publisher={Springer}
}

@article{moerland2018emotion,
  title={Emotion in reinforcement learning agents and robots: a survey},
  author={Moerland, T. M. and Broekens, J. and Jonker, C. M.},
  journal={Machine Learning},
  volume={107},
  number={2},
  pages={443--480},
  year={2018},
  publisher={Springer}
}

@article{brauwers2021general,
  title={A general survey on attention mechanisms in deep learning},
  author={Brauwers, G. and Frasincar, F.},
  journal={IEEE Transactions on Knowledge and Data Engineering},
  volume={35},
  number={4},
  pages={3279--3298},
  year={2021},
  publisher={IEEE}
}

@article{van2020brain,
  title={Brain-inspired replay for continual learning with artificial neural networks},
  author={Van de Ven, G. M. and Siegelmann, H. T. and Tolias, A. S.},
  journal={Nature Communications},
  volume={11},
  number={1},
  pages={4069},
  year={2020},
  publisher={Nature Publishing Group UK London}
}

@article{sgd2016,
  title={An overview of gradient descent optimization algorithms},
  author={Ruder, S.},
  journal={arXiv preprint arXiv:1609.04747},
  year={2016}
}

@article{assunccao2022overview,
  title={An overview of emotion in artificial intelligence},
  author={Assun{\c{c}}{\~a}o, G. and Patr{\~a}o, B. and Castelo-Branco, M. and Menezes, P.},
  journal={IEEE Transactions on Artificial Intelligence},
  volume={3},
  number={6},
  pages={867--886},
  year={2022},
  publisher={IEEE}
}

@inproceedings{landi2023cognitivenet,
  title={CognitiveNet: Enriching foundation models with emotions and awareness},
  author={Landi, R. E. and Chinnici, M. and Iovane, G.},
  booktitle={International Conference on Human-Computer Interaction},
  pages={99--118},
  year={2023},
  organization={Springer}
}

@inproceedings{yang2025,
  title={EmoEdit: Evoking emotions through image manipulation},
  author={Yang, J. and Feng, J. and Luo, W. and Lischinski, D. and Cohen-Or, D. and Huang, H.},
  booktitle={Proceedings of the Computer Vision and Pattern Recognition Conference},
  pages={24690--24699},
  year={2025}
}

@article{iovane2023smart,
  title={From Smart Sensing to consciousness: An info-structural model of computational consciousness for non-interacting agents},
  author={Iovane, G. and Landi, R. E.},
  journal={Cognitive Systems Research},
  volume={81},
  pages={93--106},
  year={2023},
  publisher={Elsevier}
}

\begin{IEEEbiographynophoto}{Riccardo Emanuele Landi} received the Master's degree in Computer Science and Engineering from Politecnico di Milano in 2021. He contributed as a research scientist in innovative startups and large companies, defining cutting-edge methodologies in computer vision and machine learning. In particular, he investigated neural network enhancement through synthetic data generation, artificial empathy, and computational cognition. He is currently Lead AI/ML Designer at Mare Group, leading the artificial intelligence and machine learning technical design and strategy. He is involved in numerous research projects in the context of machine learning for system and process optimization. Author of publications in international journals and conference proceedings, his current research interests include deep learning, affective computing, and cognitive science.
\end{IEEEbiographynophoto}

\begin{IEEEbiographynophoto}{João M. F. Rodrigues}
received the bachelor’s degree in electrical engineering in 1993, the master’s degree in systems and computer engineering in 1998, the
PhD degree in electronic and computer engineering with a speciality in computer science, in 2008, and the habilitation degree in electrical and computer engineering. He is currently serving as the vice-rector for Transfer, Innovation, and Digital University at the University of the Algarve (UAlg) in Portugal. At present is a coordinator professor in tenure with the Institute of Engineering (ISE), UAlg. Since 1994 he
has taught Curricular Units in Computer Science and Computer Vision. In addition to his academic and administrative roles, Professor Rodrigues is a member of the Research Centre NOVA LINCS -- Algarve. He has extensive research experience, having participated in more than 30 nationally or internationally funded scientific projects, several of which he served as coordinator. He has co-author more than 200 scientific papers and is a member of the editorial board of several international journals. His main areas of interest include computer vision, human-computer interaction, and human-centered Artificial Intelligence and Affective Computing, focusing on adaptive interfaces, emotion recognition, human-sense modelling, and human-machine cooperation. Details at \href{https://tinyurl.com/56fvsuwb}{https://tinyurl.com/56fvsuwb}.
\end{IEEEbiographynophoto}

\begin{IEEEbiographynophoto}{Marta Chinnici} graduated in Mathematics (2004) magna cum laude at the University of Naples (Italy), where she received her PhD in Mathematics and Computer Science (2008) with a thesis focusing on stochastic self-similar processes and application in non-linear dynamical systems. Currently, she is a Senior Researcher at ENEA in the Department of Energy Technologies and Renewable Energy Sources, ICT Division- HPC Lab, where she conducts the study, analysis, research and development on ICT with particular reference issues relating to energy efficiency in Data Center (DC), High-Performance Computing (HPC) and data science. She is a European Commission Expert as a review/evaluator in ICT and computer science for many European programs. She is a scientific responsible and manager for ENEA of many EU and national projects (PNRR-Rome Technopole); she is ENEA Representative on the DTC-Lazio Management and Coordination Committee and member of the Technical Programme Committee of various international conferences and workshops and an editor/referee and editor-in-chief of relevant journals. She is the author of books, relevant scientific articles, essays, and speeches for national and international conferences.
\end{IEEEbiographynophoto}

\end{document}